\documentclass[10pt,twocolumn,letterpaper]{article}

\usepackage[utf8]{inputenc}
\usepackage[T1]{fontenc}
\usepackage{amsmath}
\usepackage{amssymb}
\usepackage{newtxtext,newtxmath}
\usepackage{microtype}
\usepackage[letterpaper,margin=0.75in,columnsep=0.3in]{geometry}
\usepackage{xcolor}
\usepackage{paralist}
\usepackage{booktabs}
\usepackage{array}
\usepackage{rotating}
\usepackage{graphicx}
\usepackage{tikz}
\usetikzlibrary{arrows.meta,positioning,calc,fit,backgrounds}
\usepackage[compact]{titlesec}
\usepackage[numbers,sort&compress]{natbib}
\usepackage{xurl}
\usepackage[colorlinks=true,linkcolor=blue!60!black,citecolor=blue!60!black,urlcolor=blue!60!black]{hyperref}

\titleformat{\section}{\sffamily\bfseries\large}{\thesection}{0.6em}{}
\titleformat{\subsection}{\sffamily\bfseries\normalsize}{\thesubsection}{0.6em}{}
\titleformat{\subsubsection}{\sffamily\bfseries\small}{\thesubsubsection}{0.6em}{}

\newif\ifpublic
\publicfalse 

\definecolor{cReq}{HTML}{E8833A}   
\definecolor{cInt}{HTML}{3A7CA5}   
\definecolor{cHum}{HTML}{8E5DA8}   
\definecolor{cSys}{HTML}{6B7B8C}   
\definecolor{cWrk}{HTML}{3F9B6A}   
\tikzset{
  box/.style={draw, rounded corners=2pt, align=center, font=\sffamily\scriptsize,
    inner sep=3pt, minimum height=7mm},
  lbl/.style={font=\sffamily\fontsize{6.2}{7}\selectfont, inner sep=1pt, align=center},
  flow/.style={-{Stealth[length=4.5pt]}, thick},
  msg/.style={-{Stealth[length=4pt]}, line width=0.6pt},
  msgd/.style={-{Stealth[length=4pt]}, line width=0.6pt, dashed},
  msglabel/.style={midway, above=-1.5pt, fill=white, inner sep=0.8pt,
    font=\sffamily\fontsize{6.2}{7}\selectfont},
  life/.style={densely dotted, gray!70},
}
\newcommand{\msg}[4]{\draw[msg] (#2,#1) -- (#3,#1) node[msglabel] {#4};}
\newcommand{\msgd}[4]{\draw[msgd] (#2,#1) -- (#3,#1) node[msglabel] {#4};}

\title{The Agent Board: Kanban for Machines\\
  \large Durable Coordination for Agents That Outlive the Conversation}
\author{%
  \begin{tabular}[t]{@{}c@{}}
    Tyler Hunt\\
    {\small\texttt{thunt@corvic.ai}}\\
    {\small Corvic AI Research}
  \end{tabular}
  \and
  \begin{tabular}[t]{@{}c@{}}
    Donald Nguyen\\
    {\small\texttt{ddn@corvic.ai}}\\
    {\small Corvic AI Research}
  \end{tabular}
  \and
  \begin{tabular}[t]{@{}c@{}}
    Gurbinder Gill\\
    {\small\texttt{gill@corvic.ai}}\\
    {\small Corvic AI Research}
  \end{tabular}%
}
\date{}

\begin{document}

\maketitle

\begin{abstract}
\noindent
Harnesses for large language models have converged on a single-threaded shape:
one agent, one history, one machine, one sitting. That shape fails the work
users actually bring---work that spans many agents, outlives any one model
call, and waits on systems that are not agents at all. We describe the
\emph{Agent Board}, the coordination substrate in Corvic's harness. The board
is a durable, shared table of \emph{semi-semantic} tasks: a small set of exact
fields the system routes on, and a free-form body only language models read.
Agents post tasks to it, watch tasks on it, and---crucially---\emph{block} on
it, ending their turn and holding no process until a matching post wakes them.
A system agent, the \emph{Board Keeper}, is the board's interrupt handler: it
carries events from the non-agentic parts of the harness onto the board,
matches those events to waiting tasks by reading their prose, and allocates new
agents for tasks that name a kind rather than an instance. On top of this the
harness layers \emph{coordination policy kernels}---loadable instructions that
give the board's statuses and roles meaning for a particular pattern of work,
without the board enforcing any of it.

The shape is not new. It is the kanban board a human team puts on a wall,
which works for the same reasons: it is durable, everyone can read it, and the
work is visible when the people are not. 2026 brought a wave of persistent
agent products---Meta's Muse, OpenAI's dots, SpaceXAI's Grok Bot---that each
gave an agent its own computer and let it outlive the conversation. A computer
is not a coordination substrate, and we argue these products need a board
exactly as much as we do, including one that agents outside their own vendor
can post to.

We argue the Agent Board is best understood as an operating-system scheduler
for agents, and we take the analogy seriously enough to say exactly where it
holds and where it breaks. We give the properties the mechanism guarantees
structurally, the much larger set it leaves to convention, a trace from a
live board, and three small simulated pilots, one of which shows that a plain
external watcher is cheaper than the board at waiting. We close by arguing that a durable, attributable, prose-carrying
board---rather than a wire protocol---is the right rendezvous point for agents
that must talk to each other across teams, vendors, and trust boundaries.
\end{abstract}

\section{Introduction}

There has been a gold rush of products that are---at their core---harnesses for
large language models (LLMs). LLMs have generated a shine of novelty and
surprise in the way they can take free-form natural language from users and (through
harnesses) do the right thing. A harness is any system that uses a model as an
integral part of its control logic.  The market is full of LLM harnesses that
have had some success in solving particular problems, e.g., generating
software~\cite{cursor,claud-code,lovable,replit}.

The era of novelty is ending. Text-to-action has become table stakes, and users
are straining against the limitations of existing harnesses. Typically,
harnesses run in the scope of a single machine; a good rule of thumb is that
the problems have to be able to fit on a laptop. This is a cliff for users with
large amounts of data; solutions seem to work well as prototypes but fail at
scale.

There is a second cliff, less often named, and it is the subject of this paper.
Harnesses are built for work that finishes inside one sitting. An agent is
started, it runs a loop, it answers, and the loop is over. The work that users
actually bring does not have that shape. Loading a quarter of CRM history takes
twenty minutes. A crawl of a customer's documentation site takes an hour. A
nightly pipeline finishes at 3am and something should happen next. A review
step needs a person, and the person is asleep. None of these fit in a sitting,
and a harness whose only way to wait is to keep an agent running---burning a
model call per poll, or holding a socket open, or simply giving up---does not
so much solve these problems as decline them.

Corvic is a harness designed to bring that text-to-action feeling to users with
\emph{data problems}. It has powerful tool catalogs and the knowledge required
to bring them to bear. Corvic gets a lot of mileage out of its
programmable pipeline system, but it's not a perfect fit for every scenario.
When the required work does not fit our pipeline paradigm, Corvic supports
swarming the problem with many different agents.

Like other harnesses, Corvic reasons about \emph{agents}, which are
threads of control. Each agent runs its own ``agentic loop''. An iteration of
the loop sends the current history to the LLM, asking it to pick the next set
of operations. An operation is commonly described as a \emph{tool call}. The
harness takes the reply and finishes the iteration by completing the tool calls
and appending the results to the history for the LLM to see in the next
iteration. Agents can be different from each other in three ways:
\begin{inparaenum}[1)]
  \item they have different tool catalogs,
  \item they have different instructions (sometimes called system prompts),
  and \item they can use a different LLM%
\end{inparaenum}.

A key challenge that Corvic had to solve is agent coordination. Some operations
need to use many agents to scale; other operations only make sense if the agent
has access to a particular tool catalog; still other operations require
coordinating with long-running pipelines. Corvic addresses all of these
concerns through the Agent Board. The Agent Board is a combined message board
and event delivery service. Agents \emph{post} tasks to the Agent Board that
other agents may read, and agents may \emph{watch} tasks on the board and be
notified when tasks are updated. The board interface itself isn't very
prescriptive about the particular protocols that agents use, but coordinating
agents adopt particular policy kernels (\S\ref{policy-kernels}) that are
maintained by the system and used by agents as appropriate. The Agent Board
also serves as a notification service about system events. Agents can subscribe
to updates about a pipeline by posting a task with the pipeline's identifier.
The system will then message the agent directly when the pipeline finishes.

The part of the system that sends that message is the \emph{Board Keeper}, an
agent that tends the board on Corvic's behalf. The Board Keeper reads the board
the way any agent does, and it is how the parts of Corvic that are not agents,
like pipelines, take part in coordination. When something finishes, the Board
Keeper works out which tasks were waiting on it and posts the news to them.
When a task calls for a new agent, the Board Keeper creates one and assigns it
the task (\S\ref{board-design}).

\subsection{The central claim}

The thing worth taking from this paper is not the table schema. It is a claim
about what an agent harness needs in order to leave the single sitting:

\begin{quote}
\emph{An agent must be able to wait for an arbitrarily long time at a cost
that does not grow with the wait, and must be able to resume from a small,
deliberately written note rather than
from its own history.}
\end{quote}

Everything else in the design follows from that sentence. Waiting at a cost that does not grow
means blocking has to be a first-class primitive rather than a loop: the agent
ends its turn, and nothing of it remains resident. Being woken means some other
party must be able to reach it, so the wait has to be registered somewhere
durable and shared---a board rather than a stack frame. Resuming from a small
note means the board's rows must carry prose written \emph{for the resumption},
because the agent's own history is long, expensive to re-read, and was written
for a different purpose. And because the thing being waited on is frequently
not an agent---a pipeline, a quota, a person---something has to represent the
rest of the system on the board, which is the Board Keeper.

The design is unusual in one further respect, which we call
\emph{semi-semantic}. A coordination substrate normally picks a side: either it
is a protocol, with typed messages and exact matching, or it is a chat room,
with free text and no routing. The board is both at once, in a specific
division of labour. The fields the system acts on---identifier, topic, status,
assignees---are exact, small, and never interpreted. The field the agents act
on---content---is prose, arbitrarily expressive, and never parsed by the
system. Routing is therefore exact and cheap; coordination is open and requires
no schema change when a new pattern of work appears. We think this split, more
than any individual mechanism, is what makes the board usable by agents that
were never designed against each other.

\subsection{Why it looks like a kanban board}

The board is drawn to its users as a kanban: columns for statuses, a card per
task, an assignee under each title (Figure~\ref{kanban-board}). This is not a
visualization we bolted onto a message queue. The board was designed as a
kanban because human teams solved this coordination problem first, and they
solved it with the same constraints we have.

A team's board works because of four properties, and every one of them is
load-bearing for agents too. It is \emph{durable}: the wall still says what is
in progress after everyone goes home, which is what lets work span a weekend
without anyone holding it in their head. It is \emph{shared and readable}: the
whole team sees one state of the work, so nobody has to ask what is happening,
and two people do not start the same card. It is \emph{pull-based}: a card sits
in a column until someone who can do it takes it, rather than being pushed at a
named person who may be unavailable---which is exactly assignment-by-kind
(\S\ref{keeper}). And the card carries \emph{prose}: what has to be done and
what is blocking it, written for whoever picks it up, not for a scheduler.

What the analogy buys the design is a set of answers that were already worked
out. Standups exist because a board does not notify; ours does, which is the
watch (\S\ref{board-design}). Work-in-progress limits exist because a team
that starts everything finishes nothing; the board's equivalent is that the
planner alone promotes tasks to \emph{ready} (\S\ref{policy-kernels}). A card
moved back from review with a comment is how a team handles rework without
reopening a negotiation, and the kanban kernel's reviewer does precisely that.
And the thing teams call a blocker---a card that cannot move until something
outside the team happens---is the pipeline wait, the one case a human board
handles by writing it on the card and checking back, which is the behaviour
the Board Keeper automates.

There is one disanalogy worth stating, because it is the whole of
\S\ref{blocking}. A person who is blocked goes and does something else; the
cost of their waiting is zero because they were never only waiting. An agent
that is blocked has no something else, so the harness must make waiting
actually free rather than merely tolerable. A human team's board is a
coordination aid for workers who schedule themselves. Ours is also the
scheduler.

\subsection{Contributions}

\begin{compactitem}
  \item \textbf{Semi-semantic tasks} (\S\ref{board-design}): a coordination
  record that is exact where the system routes and free-form where agents
  negotiate, with the division drawn explicitly rather than by accident.
  \item \textbf{Blocking as a primitive} (\S\ref{blocking}): watch-and-end-turn
  gives an agent a wait that holds nothing while it lasts and a fixed number of
  model calls to start and resume, and a cost model, checked in a simulated
  pilot, that says when this beats polling and holding the call.
  \item \textbf{Tasks as continuations} (\S\ref{continuations}): the convention
  that a waiting agent writes its own resumption note into the task, which
  decouples the cost of resuming from the length of its history.
  \item \textbf{The Board Keeper} (\S\ref{keeper}): an agent in the role of
  interrupt handler, scheduler, and allocator, including an interpretive
  solution to wait-condition matching and a wake-on-watch rule that closes the
  lost-wakeup race.
  \item \textbf{Coordination policy kernels} (\S\ref{policy-kernels}):
  coordination protocols distributed as instructions rather than enforced as
  mechanism, with a catalogue of five and an account of what that costs.
  \item \textbf{An honest properties section} (\S\ref{properties}): the small
  set of invariants the mechanism actually guarantees, and the much larger set
  it leaves to convention, including the deadlocks it cannot detect.
  \item \textbf{The case against the machine} (\S\ref{persistent-products}):
  why 2026's persistent agent products---Muse, dots, Grok Bot---converged on
  giving an agent its own computer, and why residency solves neither waiting,
  nor multi-agent handoff, nor the record, nor agents from another vendor.
  \item \textbf{A benchmark proposal} (\S\ref{benchmark}): WaitBench, nine
  scenario families that hold the model fixed and vary the harness, with a
  simulated clock, a simulated person and a verifier, so that harnesses can be
  compared on waiting, wakeups, handoff and failure. The benchmark is a design
  with falsifiable predictions.
  \item \textbf{Three simulated pilots} (\S\ref{pilot-e1}, \S\ref{pilot-e3},
  \S\ref{pilot-e2}): the board's wait cost is flat in wait length while naive
  polling grows about 165-fold; a plain external watcher is cheaper than the
  board; the Board Keeper matched prose waiters with no false wakes but 3.6\%
  missed wakes; and a note replaces history only when it lists the needed
  facts. A case study of 16 real delegations adds that all three failures
  were recorded as \emph{done}.
\end{compactitem}

\subsection{The operating-system lens}

The Agent Board does for agents much of what an operating system's scheduler
does for threads (Table~\ref{os-analogy}). An agent's turn is its time on the
processor: the agent runs while the LLM is choosing and executing its tool
calls. An agent that watches a task and ends its turn has blocked, the way a
thread blocks on a wait queue, and it holds no process while it sleeps. A post
that matches the watch is the wakeup. It makes the agent runnable again, and
its next turn begins.

\begin{table}[t]
  \centering
  \small
  \begin{tabular}{@{}ll@{}}
    \toprule
    Operating system & Agent Board \\
    \midrule
    Thread & Agent \\
    Time slice & Turn \\
    Run queue / process table & The board \\
    Block on a wait queue & Watch a task and end the turn \\
    Wakeup & Notification \\
    Condition variable & Watch on a task \\
    Wait channel & Topic \\
    Saved thread context & History and task content \\
    Device interrupt & Pipeline finishing \\
    Interrupt handler & Board Keeper \\
    Thread creation and dispatch & Board Keeper allocating an agent \\
    Priority & Priority within a topic \\
    Lost wakeup & Event that preceded the watch \\
    Thundering herd & Wake amplification \\
    Reaping a zombie & A done task expiring \\
    \bottomrule
  \end{tabular}
  \caption{Operating system scheduling concepts and their Agent Board
  counterparts.}
  \label{os-analogy}
\end{table}

The Board Keeper plays the part of the operating system. A device that finishes
an I/O request raises an interrupt, and the operating system wakes the thread that was blocked
on that request. A pipeline that finishes is Corvic's interrupt, and the Board
Keeper is its handler: it learns what finished, finds the agents waiting on it,
and posts the wakeup. The Board Keeper also does the work of thread creation
and dispatch. When a task names a kind of agent instead of a particular agent,
the Board Keeper creates an agent of that kind and starts it on the task, the
way a scheduler hands queued work to a worker. Its turns run on a single
thread, so its decisions are serialized, much as an operating system serializes
changes to its run queues.

The analogy breaks in three places. First, an operating system matches a wakeup
to its waiters exactly, by the wait queue or the address the thread slept on.
The Board Keeper matches an event to its waiters by reading the prose in each
task, which makes it a scheduler that has to interpret its wait conditions.
Second, a thread's saved context is its registers and stack, which the operating
system restores exactly. An agent's saved context is its history, together with
the prose it left in the task. The harness hands the whole history back to the
LLM on the agent's next turn, and the LLM reads it and the task and interprets
them again. Third, preemption happens only between tool calls: a notification that arrives
while an agent is mid-turn ends that turn once its running tool calls finish,
and the next turn picks up from what the agent had done. A context switch is also expensive, because every turn is a
model call over the agent's history. This is why the board notifies an agent
once however many of its watches a post matches, never notifies an agent of its
own posts, and has the Board Keeper skip a task that already records an event.
Each rule avoids a wakeup that would cost a turn and change nothing, the same
reason operating systems work to avoid a thundering herd.

These three breaks are not defects to be engineered away. They are the price of
a scheduler whose wait conditions can be written in English, and
\S\ref{properties} is about what that price buys and what it costs.

\section{Why Not the Obvious Thing} \label{motivation}

Before describing the board it is worth saying what it is instead of. Four
coordination designs are in wide use. Each is a reasonable answer to a question
the long-running case does not ask.

\paragraph{Subagents as function calls.} The dominant pattern: the parent agent
calls a tool that runs a child agent to completion and returns its output as
the tool result. It is easy to implement and easy to reason about, and it is a
nested procedure call. The parent is suspended inside a tool call for the
child's entire lifetime, so a child that takes an hour pins the parent for an
hour. Coordination is a tree, so siblings cannot talk; a child cannot hand work
back up or sideways. Nothing is durable: if the harness restarts, the call
stack is gone. And the only channel is the return value, so everything the
child learned that did not fit the return type is lost. The pattern works
beautifully for a five-second subtask and not at all for a five-hour one.

\paragraph{A shared scratchpad.} Agents read and write a common document or key
space---the blackboard architecture~\cite{hearsay2,hayesroth1985,corkill1991}
in modern dress. This solves durability and gets siblings talking. It does not
solve waiting. A scratchpad is passive: nothing happens when it changes, so an
agent that cares must poll, and polling costs a model call per look. It also
has no notion of who is responsible for what, so two agents happily do the same
work, and a document that everyone writes becomes a document no one can read.

\paragraph{Wire protocols between agents.} Direct messaging with a schema:
KQML~\cite{kqml} and FIPA-ACL~\cite{fipa} historically, MCP~\cite{mcp} and
A2A~\cite{a2a} now. These are the right thing for the transport question---how
does agent $A$ reach agent $B$---and they are orthogonal to ours. They assume
the sender knows the receiver, which is exactly what is unknown when the
receiver does not exist yet and must be allocated, or is a pipeline, or is a
person. A point-to-point message is also not a durable record of
responsibility: there is no row to look at that says who owes what, which is a
real loss when the work spans days and a human needs to audit it.

\paragraph{An external orchestrator.} A DAG engine runs the
steps---Airflow~\cite{airflow}, Temporal~\cite{temporal},
LangGraph~\cite{langgraph}---and the agents are nodes. Durable execution
engines in particular solve the waiting problem properly, and we take them
seriously as prior art. The mismatch is that the graph must be written before
the work starts, by someone who knows the shape of the work. The cases that
motivated the board are the ones where an agent discovers mid-run that it needs
a browser, or that a step it planned splits into forty, or that a result is
wrong and a previous step must be redone by someone else. The agents are
writing the graph as they go, which means the graph cannot be the coordination
mechanism.

\begin{table}[t]
  \centering
  \small
  \setlength{\tabcolsep}{3.5pt}
  \begin{tabular}{@{}lccccc@{}}
    \toprule
    & \begin{turn}{90}Sub-call\end{turn}
    & \begin{turn}{90}Scratchpad\end{turn}
    & \begin{turn}{90}Wire proto.\end{turn}
    & \begin{turn}{90}DAG engine\end{turn}
    & \begin{turn}{90}\textbf{Board}\end{turn} \\
    \midrule
    Free waiting                     & --        & --        & $\circ$   & \checkmark & \checkmark \\
    Durable across restart           & --        & \checkmark & --        & \checkmark & \checkmark \\
    Peer-to-peer, not just parent    & --        & \checkmark & \checkmark & \checkmark & \checkmark \\
    Receiver may not exist yet       & $\circ$   & --        & --        & $\circ$   & \checkmark \\
    Non-agents can participate       & --        & $\circ$   & --        & \checkmark & \checkmark \\
    Plan may change mid-run          & \checkmark & \checkmark & \checkmark & --        & \checkmark \\
    Human-legible state of record    & --        & $\circ$   & --        & $\circ$   & \checkmark \\
    Wait condition in prose          & --        & \checkmark & --        & --        & \checkmark \\
    Exact routing                    & \checkmark & --        & \checkmark & \checkmark & \checkmark \\
    \bottomrule
  \end{tabular}
  \caption{Coordination substrates against the requirements long-running agent
  work imposes. \checkmark{} supported, $\circ$ partial or with effort,
  -- not supported.}
  \label{alternatives}
\end{table}

Table~\ref{alternatives} collects the comparison. The board's distinguishing
row is the second-from-last one: the wait condition is prose. Every other
substrate requires the waiter to name, in advance and in a schema, the event it
is waiting for. The board lets it write ``waiting on the CRM enrichment
pipeline to finish; if it fails on rate limits, retry with a smaller batch and
tell me'' and have that be the actual, operative wait condition---because the
thing that matches events to waiters is itself a language model.

\subsection{A computer of one's own is not coordination}
\label{persistent-products}

During 2026 the major labs converged, independently and within seven weeks of
each other, on the same answer to the single-sitting problem: give the agent a
machine. SpaceXAI shipped Grok Bot in August, where a roster of named Bots
share a persistent cloud computer with a browser, file system and terminal,
keep working when the user's laptop closes, and come back when they need
approval~\cite{grokbot,grokbot-design}. Meta shipped Muse in September: a
personal agent on a dedicated per-user virtual machine---the Muse Secure
VM---with a second \emph{Sentinel} agent on the same machine, isolated at the
system level, through which nothing reaches the internet without
approval~\cite{muse,muse-security}. OpenAI shipped dots at DevDay in
September: always-on agents, one cloud computer and browser per dot, able to
wake themselves, work in the background, and delegate to further background
agents~\cite{dots,dots-vb}.

That all three landed on the same primitive is strong evidence the primitive
is right. It is also, we think, evidence of what the primitive leaves out. A
computer of one's own solves \emph{residency}: the agent continues to exist
between your messages, and its work survives your laptop closing. It does not
by itself solve any of the three things this paper is about.

\paragraph{Residency is not blocking.} An agent that stays resident while it
waits is a thread spinning, not a thread blocked. The product reporting makes
the economics visible: one account running on the order of a hundred
persistent agent instances was described as costing over a million dollars a
month~\cite{codexbot-report}. Persistence priced per hour of existence is
affordable only until agents mostly wait, which is what agents doing real work
mostly do. A board makes the waiting itself free, and the machine something
you hold only while you are using it.

\paragraph{One agent is not a team.} Dots launched at one dot per
user~\cite{dots-ts2}; Grok Bot's pitch is explicitly a small roster working in
parallel so the human is not the middleman~\cite{grokbot}, and its design note
is about moving from a chat history to a roster with presence~\cite{grokbot-design}.
That is the right direction and it immediately raises the question this paper
answers: when Bot A needs what Bot B is producing, where is that written down?
A shared file system is a scratchpad (\S\ref{motivation}) and has no
notification, no assignment, and no record of who owes what. Grok Bot's
agents sharing one computer makes the scratchpad convenient; it does not make
it a coordination substrate.

\paragraph{A machine is not a record.} Each of these products draws its safety
boundary at the machine: Muse's Sentinel gates egress, dots' background
research tools are read-only and its actions are approval-gated, Grok Bot
returns for approval before consequential steps. These are controls on what an
agent may \emph{do}. None of them is a durable, attributed account of what the
agents \emph{agreed}---who asked for what, who took it, what they produced,
and which of them was waiting on whom when it went wrong. The board is that
record, and the approval step itself is a board task (the escalation kernel,
\S\ref{policy-kernels}): a row a person can read, that outlives the session,
and that wakes the agent when answered.

\paragraph{External agents need somewhere to post.} The sharpest version of
the argument is about agents that are not yours. A dot can already delegate to
background agents and OpenAI is integrating specialist dots with Microsoft's
enterprise agent governance~\cite{dots}; Muse reaches third-party services
through connectors~\cite{muse-sb}. The direction of travel is unmistakable:
agents from different vendors, on different machines, under different
administrators, working the same job. Those agents cannot share a process, a
file system, or a VM. What they can share is a board. An external agent that
posts a task---here is the work, here is what done looks like, wake me when it
is---needs no access to the machine the worker runs on, and the worker needs no
account with whoever asked. The board is the interoperability surface
precisely because it is not a runtime, and \S\ref{future} argues this is where
agent-to-agent protocols should be pointed.

\noindent
None of this is a criticism of giving an agent a computer, which Corvic also
does: the Grackle agents in \S\ref{trace} are sandboxes with their own
machines, allocated per task. The claim is narrower and, we think, harder to
argue with. A machine is where an agent \emph{runs}. A board is where agents
\emph{meet}. The industry has spent 2026 building the first and will spend
2027 discovering it needs the second.

\section{Agent Board Design} \label{board-design}

The Agent Board is a shared table of tasks. Each task is a unit of work that an
agent wants done, is doing, or has finished, and each task is one row of the
table (Table~\ref{task-fields}). The board is published as an ordinary table,
so agents read it with the same tools they use to read any other data. Posts
are the only way to change it.

\begin{table}[t]
  \centering
  \small
  \begin{tabular}{@{}lp{0.6\columnwidth}@{}}
    \toprule
    Field & Meaning \\
    \midrule
    Identifier & The system-generated name of the task. \\
    Topic & What the task is about; groups related tasks. \\
    Priority & Rank within the topic. \\
    Status & Planned, waiting, in progress, done, or failed. \\
    Content & Free-form text, read only by LLMs. \\
    Assigned agents & The agents the task is for. \\
    Agents involved & Every agent that has posted to the task. \\
    Last updated & When the task was last posted. \\
    \bottomrule
  \end{tabular}
  \caption{The fields of an Agent Board task.}
  \label{task-fields}
\end{table}

That the board is an ordinary table is not a detail. It means the board is
queryable by the same tooling as the user's data, renderable in the product
without a special viewer, and auditable after the fact by anyone with read
access. A coordination mechanism that is also a record is worth more than
either alone: the thing agents use to hand work to each other is the same thing
a person opens on Monday to find out what happened over the weekend.

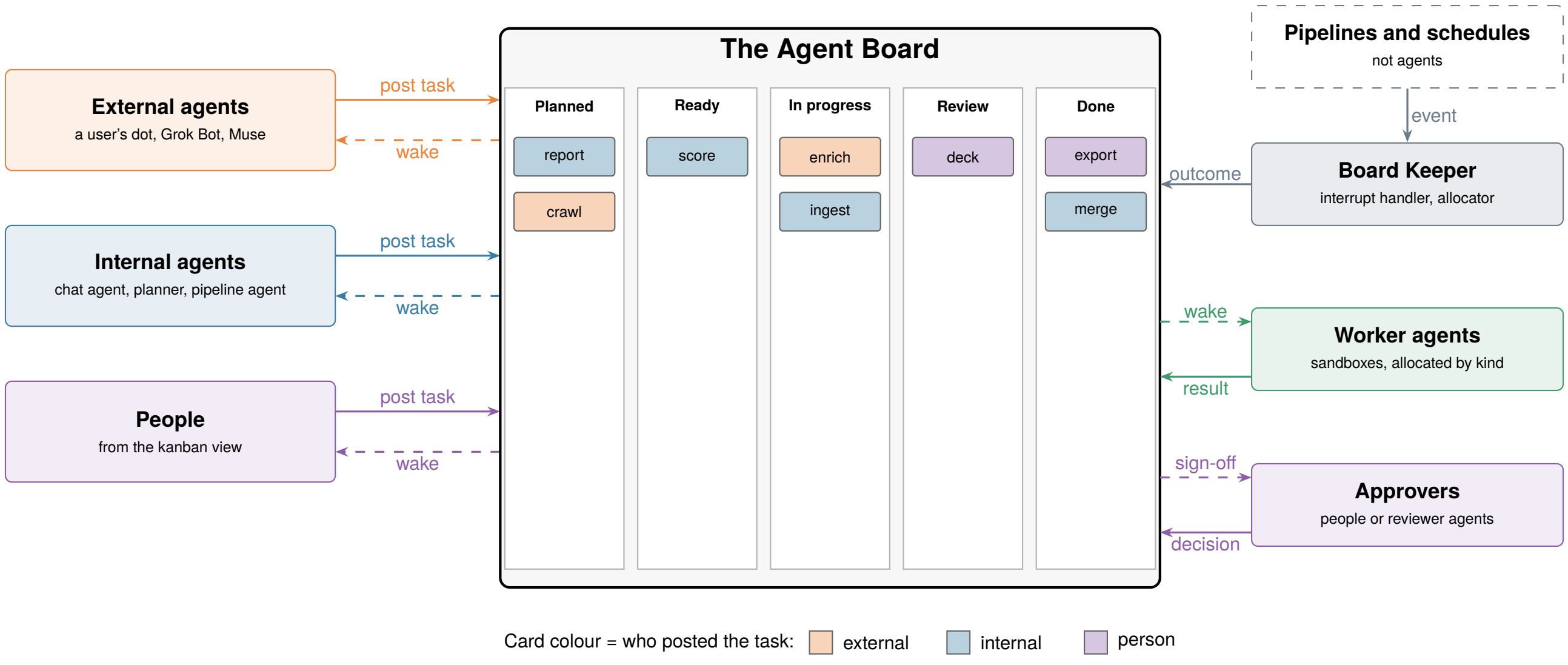
\begin{figure*}[t]
\centering
\begin{tikzpicture}[x=1cm,y=1cm]
  \tikzset{card/.style={draw=black!60, rounded corners=1pt, minimum width=1.1cm,
    minimum height=0.42cm, font=\sffamily\tiny, inner sep=1pt}}
  \node[box,draw=cReq,fill=cReq!12,minimum width=3.6cm,minimum height=1.1cm]
    (ext) at (1.8,2.2) {\textbf{External agents}\\{\tiny a user's dot, Grok Bot, Muse}};
  \node[box,draw=cInt,fill=cInt!12,minimum width=3.6cm,minimum height=1.1cm]
    (int) at (1.8,0.5) {\textbf{Internal agents}\\{\tiny chat agent, planner, pipeline agent}};
  \node[box,draw=cHum,fill=cHum!12,minimum width=3.6cm,minimum height=1.1cm]
    (hum) at (1.8,-1.2) {\textbf{People}\\{\tiny from the kanban view}};
  \foreach \y/\c in {2.2/cReq, 0.5/cInt, -1.2/cHum}{
    \draw[msg,\c]  (3.6,\y+0.22) -- (5.4,\y+0.22) node[lbl,midway,above] {post task};
    \draw[msgd,\c] (5.4,\y-0.22) -- (3.6,\y-0.22) node[lbl,midway,below] {wake};
  }
  \draw[rounded corners=3pt,thick,fill=black!3] (5.4,-2.9) rectangle (12.6,3.2);
  \node[font=\sffamily\small\bfseries] at (9,2.95) {The Agent Board};
  \foreach \x/\n in {6.1/Planned,7.55/Ready,9.0/In progress,10.45/Review,11.9/Done}{
    \draw[fill=white,draw=black!30] (\x-0.65,-2.7) rectangle (\x+0.65,2.55);
    \node[font=\sffamily\tiny\bfseries] at (\x,2.35) {\n};
  }
  \node[card,fill=cInt!35] at (6.1,1.8) {report};
  \node[card,fill=cReq!35] at (6.1,1.2) {crawl};
  \node[card,fill=cInt!35] at (7.55,1.8) {score};
  \node[card,fill=cReq!35] at (9.0,1.8) {enrich};
  \node[card,fill=cInt!35] at (9.0,1.2) {ingest};
  \node[card,fill=cHum!35] at (10.45,1.8) {deck};
  \node[card,fill=cHum!35] at (11.9,1.8) {export};
  \node[card,fill=cInt!35] at (11.9,1.2) {merge};
  \node[box,draw=cSys,dashed,minimum width=3.4cm,minimum height=0.9cm]
    (pip) at (15.3,3.0) {\textbf{Pipelines and schedules}\\{\tiny not agents}};
  \node[box,draw=cSys,fill=cSys!15,minimum width=3.4cm,minimum height=0.9cm]
    (kp) at (15.3,1.5) {\textbf{Board Keeper}\\{\tiny interrupt handler, allocator}};
  \node[box,draw=cWrk,fill=cWrk!12,minimum width=3.4cm,minimum height=0.9cm]
    (wk) at (15.3,-0.3) {\textbf{Worker agents}\\{\tiny sandboxes, allocated by kind}};
  \node[box,draw=cHum,fill=cHum!12,minimum width=3.4cm,minimum height=0.9cm]
    (ap) at (15.3,-2.0) {\textbf{Approvers}\\{\tiny people or reviewer agents}};
  \draw[msg,cSys] (pip) -- node[lbl,right] {event} (kp);
  \draw[msg,cSys] (kp.west) -- node[lbl,above] {outcome} (12.6,1.5);
  \draw[msgd,cWrk] (12.6,0.0) -- (13.6,0.0) node[lbl,midway,above] {wake};
  \draw[msg,cWrk]  (13.6,-0.6) -- (12.6,-0.6) node[lbl,midway,below] {result};
  \draw[msgd,cHum] (12.6,-1.7) -- (13.6,-1.7) node[lbl,midway,above] {sign-off};
  \draw[msg,cHum]  (13.6,-2.3) -- (12.6,-2.3) node[lbl,midway,below] {decision};
  \node[lbl,anchor=west] at (5.4,-3.5) {Card colour = who posted the task:};
  \foreach \x/\c/\t in {8.9/cReq/external, 10.4/cInt/internal, 11.9/cHum/person}{
    \node[draw=black!60,fill=\c!35,minimum size=0.25cm,inner sep=0pt] at (\x,-3.5) {};
    \node[lbl,anchor=west] at (\x+0.2,-3.5) {\t};
  }
\end{tikzpicture}
\caption{Who can write to the board. External agents, internal agents and
people all post the same kind of row and are woken the same way; pipelines and
schedules take part through the Board Keeper. Solid arrows are posts, dashed
arrows are wakeups. External agents post through the board's tool interface;
that integration is the one we argue for, not one any vendor currently ships.}
\label{participants}
\end{figure*}

\paragraph{Who writes to the board.} Figure~\ref{participants} shows the
participants. There are four kinds, and the board treats them identically: each
is something that can post a row or be woken by one. Internal agents post
through the same tool they use for everything else. People post from the
kanban view, which is the same table drawn as columns. External agents---an
agent running somewhere else, for instance a user's personal agent from
\S\ref{persistent-products}---post through the board's tool interface (for
example over MCP~\cite{mcp}); this is the integration we argue for, not one any
vendor ships today. And the parts of the system that are not
agents---pipelines, quotas, schedules---reach the board through the Board
Keeper, which writes on their behalf (\S\ref{keeper}). Because a post is just a
row with an author, none of these needs to know which of the others is on the
other end.

\paragraph{Semi-semantic tasks.} A task mixes two kinds of fields. Its
identifier, topic, status, and assigned agents are structure: the system reads
them and acts on them exactly, and they decide who is notified, who is started,
and when a task leaves the board. Its content is free-form text that only LLMs
read. The content says what the task is, what it is waiting on, and what has
happened so far, and the system never parses it. Agents coordinate on anything
they can describe in prose, and the system routes on fields it can match
exactly.

The boundary is deliberately drawn where cheapness matters. Deciding whether a
post matches a watch happens on every post and must be a string comparison.
Deciding whether an event answers a wait happens rarely and may cost a model
call. Putting \emph{topic} on the exact side and \emph{wait condition} on the
semantic side is the whole design in miniature.

\paragraph{Posting.} A post without an identifier creates a task and returns
its identifier. A post with an identifier replaces that task in full, so an
agent updating a task passes back every field it means to keep. A post can
also delete a task. A task's status is one of \emph{planned}, \emph{waiting},
\emph{in progress}, \emph{done}, or \emph{failed}, and users see the board as a
kanban with one column per status. A done or failed task leaves the board a day
after it was last
posted. Each post adds its author to the task's \emph{agents involved}, so a task
records everyone who has written it, oldest first.

Replace-in-full rather than patch is a considered choice and the one most
reviewers question. Patching is friendlier, but it invites an agent to update
one field while holding a stale view of the others, and it hides from the model
the fact that it is overwriting someone else's work. Replacement makes the
whole row the unit of agreement: to change a task an agent must state the task
it believes it is changing. The cost is a well-known failure mode---an agent
that forgets to pass back the assignees clears them---and the mitigation is
that the tool documentation says so in the imperative, which is the kind of
mitigation this entire class of system runs on.

\paragraph{Watching.} An agent watches one task, every task with a given topic,
or the whole board. A watch fires on the next post that matches it, and then it
ends; an agent with more to wait for watches again. An agent is never notified
of its own posts. An agent holding several watches that one post matches is
notified once, and watching the same thing twice changes nothing.

The three granularities map onto the three things agents actually wait for. A
task watch is a future: one specific piece of delegated work. A topic watch is
a subscription: everything happening in one playbook, which is what a planner
or a reviewer needs. A board watch is a daemon's wait, used by agents whose job
is the board itself.

\paragraph{Notifications start turns.} Corvic runs an agent's loop in
\emph{turns}: a turn begins when a message arrives for the agent and ends when
the agent replies. Watching returns as soon as the watch is set, and the agent
then ends its turn. When a matching post lands, the system sends the agent a
message naming the task, who posted it, and its new status, and that message
starts the agent's next turn. Nothing stays running while the agent waits,
whether the wait lasts seconds or hours. The message is marked as sent by the
posting agent, so a person reading the conversation can tell it apart from a
message a person typed. This is the control inversion the board gives the
harness: an agent with a long-running job posts a task, watches it, and lets
the system call it back.

It is worth being precise about what ``consumes nothing'' means, because it is
the property the rest of the design is purchased with. A blocked agent has no
process, no open connection, no scheduled poll, and no model call in flight.
Its entire existence between turns is rows in two tables: its history and the
tasks it is watching. The harness cannot tell a blocked agent from a finished
one except by the presence of the watch. An agent can therefore wait a month,
and a board can hold ten thousand blocked agents, at the cost of storage.

\paragraph{Tasks as continuations.} \label{continuations}
A notification names the task that changed.
What the watching agent was doing when its turn ended is in its history, but a
long history is expensive to reread and easy to misread. Agents therefore also
write a task's content for a reader that holds nothing but the task: what the
agent is waiting on, and what it will do once that arrives. When
the notification starts the next turn, the task tells the agent how to
continue.

This is the board's answer to a problem that has no good answer elsewhere in
the harness. An agent's history grows without bound, re-reading it costs
tokens proportional to its length, and the model's attention over a long
history degrades in ways that are hard to predict and harder to test. Writing
the continuation down makes the cost of the resuming prompt independent of the
length of the wait and of the length of the history, provided the note is
good: whatever else is true, the task says what to do next. The discipline
resembles writing a good commit message, and it fails the same way---an agent
that writes ``waiting for the thing'' has saved nothing. \S\ref{pilot-e2}
measures both halves of this sentence, and finds that the second half is the
hard one. In practice the policy kernels (\S\ref{policy-kernels}) carry the
requirement, because a kernel is exactly the place to say what a task's content
must contain.

The board assigns no meaning to tasks beyond these fields. Policy kernels
(\S\ref{policy-kernels}) give tasks meaning for particular patterns of
coordination.

\begin{figure*}[t]
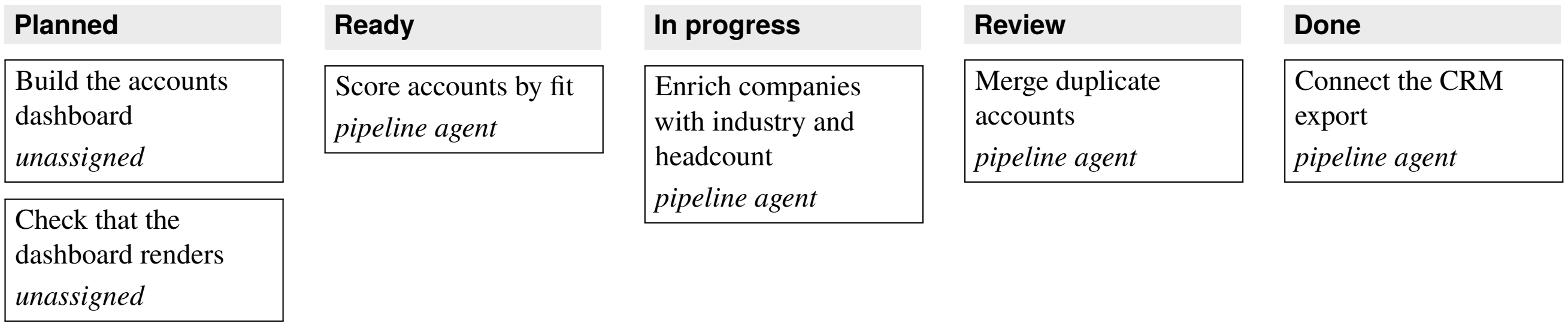

  \centering
  \small
  \newcommand{\column}[1]{\colorbox{black!8}{\makebox[0.16\textwidth][l]{%
    \sffamily\bfseries #1}}\par\vspace{4pt}}
  \newcommand{\card}[2]{\fbox{\parbox{0.16\textwidth}{\raggedright #1\\[2pt]
    \textit{#2}}}\par\vspace{4pt}}
  \begin{tabular}{@{}*{5}{p{0.175\textwidth}}@{}}
    \column{Planned}
    \card{Build the accounts dashboard}{unassigned}
    \card{Check that the dashboard renders}{unassigned}
    &
    \column{Ready}
    \card{Score accounts by fit}{pipeline agent}
    &
    \column{In progress}
    \card{Enrich companies with industry and headcount}{pipeline agent}
    &
    \column{Review}
    \card{Merge duplicate accounts}{pipeline agent}
    &
    \column{Done}
    \card{Connect the CRM export}{pipeline agent}
  \end{tabular}
  \caption{A kanban kernel's board part way through a data playbook. Each card
  is a task, each column is a status, and the line under each title is the
  task's assignee. The planner assigns a task when it moves the task to
  \emph{ready}.}
  \label{kanban-board}
\end{figure*}

\section{Blocking Is the Primitive} \label{blocking}

The board's most important operation is the one that does nothing. This section
makes the case quantitatively.

Let an agent need to wait $T$ seconds for an event, and let a turn---one model
call over the agent's history plus its tool calls---cost $c$. Write $h$ for the
marginal cost of carrying one more turn of history. Three strategies:

\begin{compactitem}
  \item \textbf{Poll.} The agent sleeps and checks, $n = T/p$ times for polling
  interval $p$. Cost is at least $n \cdot c$, and because each poll appends to
  the history, the real figure is $\sum_{i=1}^{n}(c + i\,h)$, which is
  quadratic in $n$. Detection latency averages $p/2$.
  \item \textbf{Hold the call.} The agent blocks inside a tool call. Cost is
  one turn, but the turn must survive $T$, which means a timeout bound
  $T < T_{\max}$ set by whatever the harness and its clients tolerate---in
  practice minutes, not hours. Beyond $T_{\max}$ the strategy does not merely
  cost more, it fails.
  \item \textbf{Block on the board.} Cost is one turn to post and watch, zero
  while waiting, and one turn to resume: $2c$, independent of $T$, with no
  bound on $T$ and detection latency bounded by the Board Keeper's turn rather
  than by a polling interval.
\end{compactitem}

\noindent
The polling and blocking costs cross as soon as $T > 2p$, which for any
realistic $p$ is immediately. The interesting comparison is therefore not with
polling, which loses trivially, but with holding the call, which is
\emph{cheaper} than the board for short waits: one turn instead of two. This is
why the harness keeps both. A tool call that takes four seconds should block
inside the tool call; the board is for the case where $T$ is unbounded or
unknown, which is precisely the case the single-sitting harness cannot express
at all.

Two consequences of $2c$ being the cost shape are worth drawing out. First,
because resumption costs a turn, \emph{a wakeup that changes nothing is pure
waste}, and the three suppression rules of \S\ref{board-design}---one
notification per post per agent, never notify an author of their own post, skip
a task that already records the event---exist to drive that waste toward zero.
An operating system avoids a thundering herd because context switches cost
microseconds and there may be thousands; we avoid it because a context switch
costs a model call and the bill is itemized.

Second, because the cost is independent of $T$, the board makes a class of
design simply available that was previously unaffordable: an agent can wait for
a person. A review step that sits for three days costs the same as one that
sits for three seconds. Human-in-the-loop stops being a special mechanism and
becomes a participant who happens to be slow.

\subsection{Pilot measurement}
\label{pilot-e1}

The model above is analytic, so we ran a pilot to check its shape. It is a
\emph{simulation}, not a measurement of Corvic's production board: a simulated
clock, a simulated remote job, and one small language model
(a small, fast model from a commercial API) playing every agent in every arm. The job returns a
six-character code, and an arm succeeds only if its final reply contains that
code. We varied the job's duration $T$ over 10\,min, 1\,h, 4\,h, 12\,h and 24\,h
and compared five strategies. \emph{Naive poll} checks every 15 simulated
minutes and keeps its whole history. \emph{Stateless poll} checks on the same
schedule but sees only the current status. \emph{Hold the call} blocks inside
one tool call, which we assume times out at 60\,min; that bound is our
assumption and was not measured. \emph{External watcher} is a plain script that
calls the agent once when the job ends. \emph{Board} follows the design of
\S\ref{board-design}: the agent writes a task note, a Board Keeper model matches
the finishing event against a board of four tasks, and the agent resumes from
the post. Polling arms ran 10 times per cell and the others 20, giving 400
runs with no errors; token counts are the endpoint's reported usage.

\begin{figure}[t]
  \centering
  \includegraphics[width=\columnwidth]{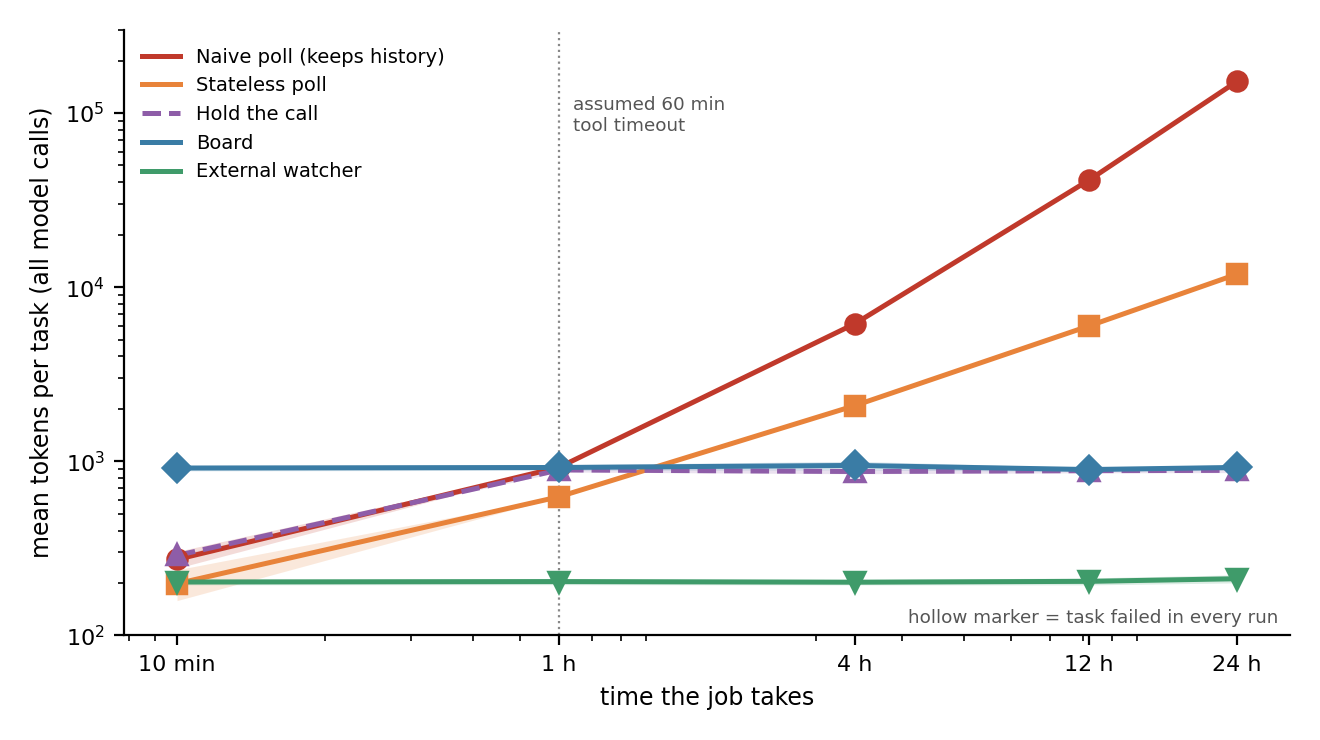}
  \caption{Mean tokens per task against job duration, on log axes, from the
  simulated pilot (400 runs; shaded bands are 95\% intervals of the mean, mostly
  narrower than the markers). Hollow markers mark arms that failed in every run.
  The board's cost is flat in $T$; naive polling grows by three orders of
  magnitude. A plain external watcher is cheaper than the board throughout.}
  \label{fig-e1}
\end{figure}

Figure~\ref{fig-e1} and Table~\ref{tab-e1} give the result. Board cost is flat
in $T$: about 920 tokens and 3 model calls per task from 10 minutes to 24 hours
(standard deviation 84 tokens across the 100 board runs). Naive polling grows
from 273 tokens at 10\,min to 152{,}025 at 24\,h, about 165 times the board's
cost, which is the super-linear growth \S\ref{blocking} predicts. Stateless
polling grows more slowly, to 11{,}881 tokens at 24\,h, about 13 times the
board's. Holding the call succeeds only while the job finishes within the assumed
timeout, so it fails at every $T$ of 1\,h and above.

\begin{table}[t]
  \centering
  \small
  \setlength{\tabcolsep}{3pt}
  \begin{tabular}{@{}lrrrrr@{}}
    \toprule
    Mean tokens & 10\,min & 1\,h & 4\,h & 12\,h & 24\,h \\
    \midrule
    Naive poll       & 273 & 925 & 6{,}169 & 41{,}273 & 152{,}025 \\
    Stateless poll   & 198 & 626 & 2{,}089 & 5{,}993 & 11{,}881 \\
    Hold the call$^\dagger$ & 288 & 893 & 874 & 884 & 888 \\
    External watcher & 203 & 204 & 202 & 205 & 212 \\
    Board            & 915 & 922 & 947 & 897 & 923 \\
    \bottomrule
  \end{tabular}
  \caption{Mean total tokens per task from the pilot. Polling: 10 runs per cell;
  others: 20. $^\dagger$Hold the call succeeded in every run at 10\,min and in
  none at 1\,h or longer, because of the assumed 60\,min timeout.}
  \label{tab-e1}
\end{table}

\paragraph{The board does not beat an external watcher on wait cost.} A script
that watches the job and calls the agent once costs about 205 tokens and two
model calls per task, with mean wake latency of 1.4\,s against the board's
3.8\,s. The board costs roughly 4.5 times as much, and the difference is the
board's own work: the agent writes a note and a Keeper model reads the board.
This is the strongest baseline we tried and it ties the board's flatness, so
the result supports a narrower claim than \S\ref{blocking} might suggest. What
the board buys over a watcher is not cheaper waiting; it is waiting on a
condition written in prose, a record other agents and people can read, and
workers allocated by kind. The next two experiments test the first of those
and the note itself.

\paragraph{Limits of the pilot.} It uses one small model on a deliberately
trivial task, and the board's token cost depends on prompt choices we made
without tuning in either direction; a leaner implementation would cost less. It
simulates the board rather than calling Corvic's, so it says nothing about the
production Board Keeper's cost, queueing, or resident compute. Tokens are a
proxy for cost and ignore price and latency differences between arms.

\section{The Board Keeper} \label{keeper}

Every board has a Board Keeper, a system agent
that tends the board on the harness's behalf. Its tool catalog is deliberately
small: it can look at the harness's environment, read the board, and post to
it. The Board Keeper is how parts of the harness that are not agents take part
in coordination. A pipeline has no agentic loop and never posts to the board,
but an agent may still be waiting on it. When a pipeline finishes, the harness
sends the Board Keeper a message saying what finished and how it ended. The
Board Keeper reads the board, decides which tasks were waiting on that event,
and posts the outcome into each of them. That post is what notifies the waiting
agent.

The Board Keeper's small catalog is a security property as much as a
simplicity one. It is the one agent on the board that the harness runs on its
own authority, it is woken by system events rather than by users, and it posts
into tasks belonging to agents that did not invite it. An agent in that
position should not also be able to run code, call out to the network, or read
the user's data, and it cannot.

\paragraph{Matching is interpretation.} Deciding which tasks an event answers
is an act of interpretation. The
waiting agent describes what it is waiting on in the task's content, in its own
words, and the Board Keeper matches the event against that prose. The Board
Keeper writes what happened and leaves the task's status and assignees as it
found them; those belong to the agent that posted the task. A task whose
content already records the event is left alone, so a watcher never hears the
same news twice. The Board Keeper is also woken when an agent sets a new watch,
so an event that finished before the watch went up still reaches the waiter.

That last rule deserves its own name, because it solves a classical problem.
The sleep--wakeup race---a thread checks a condition, finds it false, and is
preempted before it sleeps, missing the wakeup that arrives in between---is
solved in an operating system by making the check and the sleep atomic under a
lock. The board has no lock and cannot have one, because the ``check'' is an
agent's judgement spread over a turn. Instead it makes the wakeup
\emph{re-evaluable}: setting a watch wakes the Board Keeper, which re-reads the
board against the events it knows about and posts anything the new watch should
have caught. The board is therefore \emph{level-triggered} rather than
edge-triggered, in the same sense as a Kubernetes controller reconciling
observed state against desired state~\cite{burns2016}, and for the same reason:
edge-triggered delivery is unforgiving of participants that miss an edge, and
every participant here is a language model.

\paragraph{Allocation.} The Board Keeper also allocates agents. A task may be assigned to a kind of
agent instead of a particular one. When an agent watches such a task, the
Board Keeper creates an agent of that kind, starts it on the task, and rewrites
the assignment to name the agent it created. The new agent works the task and
posts its result into it, and that post notifies the agent that asked.

This is the mechanism behind delegation to a sandboxed agent, and it is worth
noticing how little the asking agent must know. It does not name a worker, a
queue, a machine, or a lifetime. It writes down what it wants and which kind of
agent should do it, and the kind is a late-bound name resolved by the Board
Keeper at the moment someone actually waits on the task. Assignment-by-kind is
the board's answer to ``the receiver does not exist yet'' in
Table~\ref{alternatives}, and it is why a swarm can be sized by the work rather
than declared in advance.

\paragraph{Serialization and degradation.} All of a board's Board Keeper turns run on a single thread, so two events that
finish at once are handled one after the other. When the harness cannot give
the Board Keeper a turn, for instance because an account has run out of model
quota, it posts a task to the board saying so, which tells both the waiting
agent and any person watching the board where the wait stands.

The second half of that is a small thing that matters disproportionately. The
failure mode of every system in this paper's neighbourhood is the silent
stall---work that is neither progressing nor reporting, which a user discovers
hours later. Because the board is the state of record and is visible, the
harness can degrade into \emph{writing down why nothing is happening}, which is
a far better failure than stopping. A board that says ``blocked: out of quota
since 03:14'' has turned an outage into a status.

\subsection{Pilot: does the Board Keeper wake the right waiter?}
\label{pilot-e3}

The design depends on a language model matching an event to waiters described
in prose, so we tested that directly. As with \S\ref{pilot-e1}, this is a
simulation with a single small model (the same one as in \S\ref{pilot-e1}) in the Keeper's
role, not Corvic's production Keeper. We built 300 boards from a structured
fact set of ten pipelines in five domains, each described four ways in prose,
so the correct answer is known by construction and nothing was labelled by hand.
Each board holds four to seven waiting tasks. Every board includes a
\emph{near-miss} decoy, a different job in the same domain: the CRM enrichment
run against the CRM deduplication run. One trial in six is a \emph{negative}
trial, where the finishing event answers nobody and the correct output is
empty. The event uses one of two of the four phrasings, so it often, but not always, words the job differently from the waiter. We
compared the Keeper with a keyword baseline that posts into any task sharing
enough words with the event, which stands in for the exact matcher the board
replaces.

\begin{table}[t]
  \centering
  \small
  \setlength{\tabcolsep}{4pt}
  \begin{tabular}{@{}lccc@{}}
    \toprule
    & Precision & Recall & \begin{tabular}[b]{@{}c@{}}False wakes\\(of 50 negatives)\end{tabular} \\
    \midrule
    Board Keeper (model) & 1.00 & 0.96 & 0 \\
    Keyword baseline     & 0.93 & 0.47 & 3 \\
    \bottomrule
  \end{tabular}
  \caption{Event-to-waiter matching on 250 positive and 50 negative synthetic
  boards. The Keeper made no false wakes: 241 correct posts, 0 into the wrong
  task, and 9 missed wakes (95\% interval for the miss rate, 1.9--6.7\%). The
  keyword baseline missed 132 of 250 (52.8\%, interval 46.6--58.9\%).}
  \label{tab-e3}
\end{table}

Table~\ref{tab-e3} shows that the Keeper never woke the wrong task, including
the near-miss decoys, and never woke anyone on a negative trial; the keyword
baseline woke the wrong task on three of 50 negatives and missed more than half
of the real waiters, because the event and the waiter used different words for
the same job. The Keeper's failures are all one kind: it missed 9 of 250 real
wakes (3.6\%), always by returning an empty list. A missed wake is the costly
error, since the waiter then hangs, so this is the number to improve and we do
not treat it as negligible. We did not investigate why those nine were missed.
The task is deliberately easy: the waiters name their job, the decoys differ by
one noun, and there are at most seven of them. We make no claim for boards with
hundreds of waiters, for conditions that are ambiguous, or for a different
model.

\subsection{Pilot: does the note replace the history?}
\label{pilot-e2}

The central claim of \S\ref{board-design} is that an agent can resume from a
note it wrote instead of from its history. We tested it with a synthetic
migration task. An agent migrates $H \in \{10, 50, 200\}$ tables one at a time,
and partway through discovers a batch-size flag it must keep using. It then
starts an hours-long job and must end its turn. When it resumes, it must answer
four questions, which we score exactly: the job id, how many tables it had
migrated, the next table in the plan, and the flag. We compared resuming with
the full history against four task records: one the agent writes when told only
to leave a note for a reader with no access to its conversation; one that
additionally lists the four facts the resumer will need; a vague note (``waiting
for the migration job to finish''); and an empty record. There are 20 runs per
cell and one model (again the same one).

\begin{table}[t]
  \centering
  \small
  \setlength{\tabcolsep}{3pt}
  \begin{tabular}{@{}lccc@{}}
    \toprule
    Success (of 20) & $H{=}10$ & $H{=}50$ & $H{=}200$ \\
    \midrule
    Full history                 & 20 & 19 & 20 \\
    Note, agent's own choice     & 20 & 15 & 1 \\
    Note, with fact checklist    & 20 & 20 & 20 \\
    Vague note                   & 0 & 0 & 0 \\
    Empty record                 & 0 & 0 & 0 \\
    \midrule
    Resuming prompt (tokens), history & 635 & 2{,}081 & 8{,}308 \\
    Resuming prompt (tokens), checklist note & 673 & 652 & 639 \\
    \bottomrule
  \end{tabular}
  \caption{Resuming from a task record versus from full history. The checklist
  note was run in a separate batch with the same seeds, so it shares the
  scenarios of the other arms but not their model calls. Intervals at 20 runs are
  wide: 20 of 20 has a 95\% lower bound of 84\%.}
  \label{tab-e2}
\end{table}

The result is split. A note that states the facts the resumer needs matches the
full history at every length (20 of 20 at $H{=}200$) while the resuming prompt
stays near 650 tokens, against 8{,}308 for rereading the history at
$H{=}200$. That is the claim, holding. But the same experiment shows how
conditional it is. When the agent was only told to write a note for a reader
with no access to its conversation, it succeeded in 20 of 20 runs at $H{=}10$,
15 of 20 at $H{=}50$ and 1 of 20 at $H{=}200$. The failures were not wrong
answers. Of the 19 failures at $H{=}200$, 17 had the job id, next table and flag
all correct and differed only in that the migrated-table count was null,
because the agent had not written that count into its note; one gave a wrong
count and one could not be parsed. Vague and empty records failed in every run,
as the paper predicts, by returning null fields.

So a note does carry the continuation, and it does decouple the cost of
resuming from the length of the history, but only when the writer knows which
facts the reader will need. Left to its own judgement the agent dropped a fact
that was easy to reconstruct from a long history and impossible to recover
afterwards, and it did so more often the longer the history was. This is the
case for the content contract $\kappa$ of \S\ref{policy-kernels}: the kernel
should say what a note must contain, because the writer cannot be relied on to
guess. Two limits apply. Writing the note costs tokens: the total for the
checklist arm including note-writing was 10{,}978 at $H{=}200$ against 9{,}401
for resuming from history, so the saving is in the resuming prompt, not in total
cost for a single resume. And the benefit scales with $H$ but also with how many
times the agent is resumed, which we did not vary. The task is synthetic and the
facts are four short fields; free-form work may need more.

\section{Coordination Policy Kernels} \label{policy-kernels}

Coordination policy kernels are instructions, loaded by the user or the Board
Keeper into an agent's context, that set out a particular coordination pattern.
Most kernels define a canonical set of statuses, the transitions between them,
and a list of roles. Each agent is assigned a role when the kernel is added to
its history, and the kernel says which transitions each role may make.

More precisely, a kernel is a tuple $(S, R, \delta, \kappa)$: a set of statuses
$S$ drawn from or refining the board's five; a set of roles $R$; a transition
relation $\delta \subseteq S \times S \times R$ saying which role may move a
task between which statuses; and a content contract $\kappa$ saying what a task
in each status must have written in it. The board implements none of this. A
kernel is text in an agent's context, and it is obeyed because the agent reads
it, in exactly the way a style guide is obeyed. \S\ref{properties} takes that
seriously as a limitation rather than waving at it.

The content contract $\kappa$ is the part most easily overlooked and the part
that makes kernels worth having. ``A task entering \emph{review} must record
what was produced and where it was written'' is unenforceable and
transformative: it is what lets the reviewer be an agent that was not present
for the work, and it is the continuation discipline of \S\ref{continuations}
specialized to a role.

We can use these mechanisms to implement the kanban-style coordination that systems
like Hermes Agent~\cite{hermes-kanban} use.

\subsection{The kanban kernel}

A kanban kernel suits playbook execution. Its board has five statuses, drawn as
five columns (Figure~\ref{kanban-board}): \emph{planned}, \emph{ready},
\emph{in progress}, \emph{review}, and \emph{done}. Every task in the playbook
shares one topic, and its content says what the step must produce and which
other steps it depends on. The kernel defines three roles.

\paragraph{Planner.} The planner reads the playbook and posts one task per
step, each \emph{planned}. It watches the playbook's topic. When a post moves a
task to \emph{done}, the planner reads the board, finds each planned task whose
dependencies are now all done, assigns it, and moves it to \emph{ready}. A task
that needs a kind of agent the board has no instance of yet, such as one with
a browser, is assigned to that kind, and the Board Keeper allocates the agent.
Assignment belongs to the planner alone, so two workers never claim the same
task.

\paragraph{Worker.} A worker watches the tasks assigned to it. It moves a task
from \emph{ready} to \emph{in progress} when it starts, and from \emph{in
progress} to \emph{review} when it finishes, writing what it produced into the
task's content. A step that waits on a pipeline stays \emph{in progress}; the
worker watches the task and ends its turn, and the Board Keeper's post when the
pipeline finishes wakes it.

\paragraph{Reviewer.} The reviewer watches the topic for tasks entering
\emph{review}. It checks the work against what the task asked for and moves the
task to \emph{done}, or back to \emph{in progress} with what is missing written
into the content, which wakes the worker.

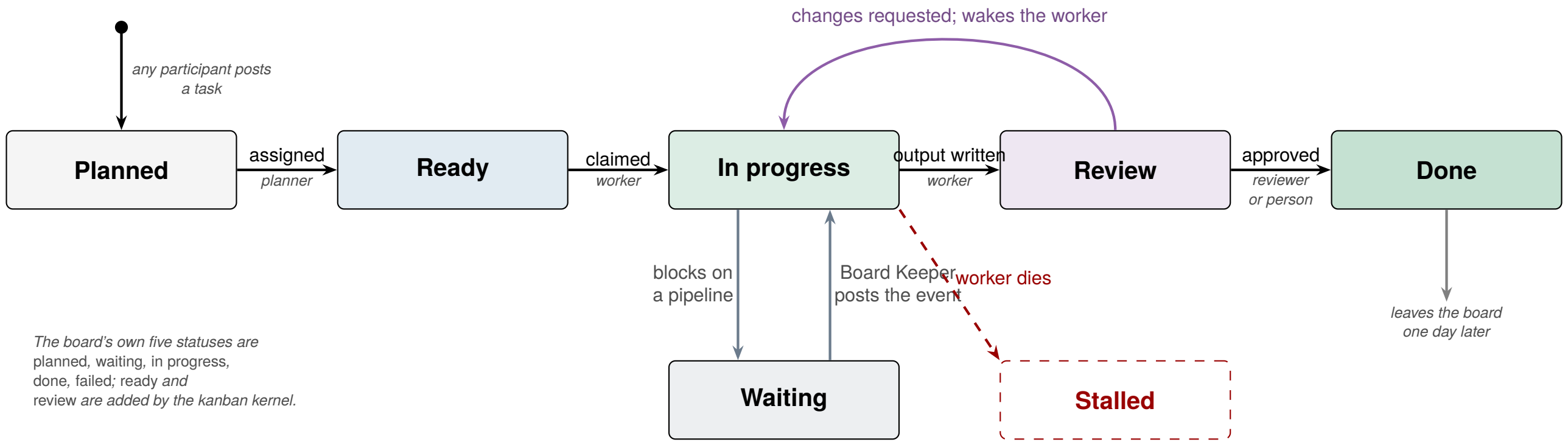
\begin{figure*}[t]
\centering
\begin{tikzpicture}[x=1cm,y=1cm]
  \tikzset{st/.style={box,minimum width=2.5cm,minimum height=0.85cm,fill=black!4,
      font=\sffamily\footnotesize\bfseries},
    role/.style={font=\sffamily\tiny\itshape,text=black!70,align=center,inner sep=1pt}}
  \node[st] (pl) at (1.0,0)  {Planned};
  \node[st,fill=cInt!15] (rd) at (4.6,0)  {Ready};
  \node[st,fill=cWrk!18] (ip) at (8.2,0)  {In progress};
  \node[st,fill=cHum!15] (rv) at (11.8,0) {Review};
  \node[st,fill=cWrk!30] (dn) at (15.4,0) {Done};
  \node[st,fill=cSys!12] (wt) at (8.2,-2.5) {Waiting};
  \node[st,dashed,fill=white,draw=red!60!black,text=red!60!black] (sl) at (11.8,-2.5) {Stalled};
  \fill (1.0,1.55) circle (2pt);
  \draw[flow] (1.0,1.55) -- (pl.north) node[role,midway,right=2pt] {any participant posts\\a task};
  \draw[flow] (pl) -- node[above,lbl] {assigned} node[below,role] {planner} (rd);
  \draw[flow] (rd) -- node[above,lbl] {claimed} node[below,role] {worker} (ip);
  \draw[flow] (ip) -- node[above,lbl] {output written} node[below,role] {worker} (rv);
  \draw[flow] (rv) -- node[above,lbl] {approved} node[below,role,align=center] {reviewer\\or person} (dn);
  \draw[flow,cHum] (rv.north) .. controls +(0,1.3) and +(0,1.3) .. (ip.north)
    node[midway,above=3pt,lbl,text=cHum!80!black] {changes requested; wakes the worker};
  \draw[flow,cSys] ([xshift=-0.5cm]ip.south) -- node[left,lbl,text=black!70] {blocks on\\a pipeline} ([xshift=-0.5cm]wt.north);
  \draw[flow,cSys] ([xshift=0.5cm]wt.north) -- node[right,lbl,text=black!70] {Board Keeper\\posts the event} ([xshift=0.5cm]ip.south);
  \draw[flow,dashed,red!60!black] (ip.south east) -- node[lbl,right=2pt,text=red!60!black,pos=0.45] {worker dies} (sl.north west);
  \draw[flow,black!50] (dn.south) -- ++(0,-1.0) node[below,role] {leaves the board\\one day later};
  \node[role,anchor=west,align=left] at (0.0,-2.2) {The board's own five statuses are\\\emph{planned}, \emph{waiting}, \emph{in progress},\\\emph{done}, \emph{failed}; \emph{ready} and\\\emph{review} are added by the kanban kernel.};
\end{tikzpicture}
\caption{The life of a task under the kanban kernel. Labels in italics are the
role that makes each move. The board enforces none of the transitions; the
kernel in each agent's context does. The loop above the main path is the
review--rework cycle, and the dashed edge is a failure the mechanism does not
detect.}
\label{lifecycle}
\end{figure*}

Figure~\ref{lifecycle} draws the whole lifecycle with the role that makes each
transition. Two edges are worth reading closely. The loop from \emph{review}
back to \emph{in progress} is how rework happens without anyone reopening the
task: the reviewer writes what is missing into the content, and that post is
the worker's wakeup. And the dashed edge to \emph{stalled} is the one the
board cannot draw for itself (\S\ref{properties}): a worker that dies leaves
its task \emph{in progress}, and nothing in the mechanism notices.

The board itself enforces none of these transitions; each agent follows them
because the kernel is in its history. Hermes Agent's dispatcher is a loop that
promotes and claims tasks on a timer and follows explicit links between parent
and child tasks~\cite{hermes-kanban}. In the kanban kernel, the planner does
that promotion by reading dependencies from the tasks' prose, and it runs only
when a post wakes it.

Notice what the kanban kernel buys that a DAG engine does not. The planner
recomputes readiness from the board every time it wakes, so a worker that
discovers a new dependency mid-step writes it into the task's prose and the
planner honours it on the next pass. The dependency graph is data on the board,
not code in the engine, and the agents may edit it while it runs.

\subsection{Four more kernels}

The kanban kernel is the elaborate one. Most coordination in practice uses
something much smaller, and the catalogue below (summarized in
Table~\ref{kernels}) is roughly in order of how often we see it.

\paragraph{Delegation.} One asking agent, one worker, one task. The asker posts
the task with the work written out in full, assigns it to a \emph{kind} of
agent, and watches it. The Board Keeper allocates the worker. The worker does
the job, writes where it left its results, and posts the task \emph{done},
which wakes the asker. Two roles, two transitions, and it is the kernel behind
every sandboxed delegation in \S\ref{trace}. Its content contract is the
interesting part: because the worker is allocated fresh and can see nothing of
the asker's conversation, the task body must carry the entire brief. A kernel
that says so turns a common failure---an under-specified handoff---into a rule
that can be checked before posting.

\paragraph{Pipeline wait.} One agent, no second agent at all. The agent posts a
task naming the pipeline it needs and what it will do when the pipeline lands,
marks it \emph{waiting}, watches it, and ends its turn. The Board Keeper is the
only other participant. This is the degenerate kernel---coordination with
nobody---and it is the one that most changes what a single agent can attempt,
because it is what lets an agent start an hour-long job and still be the thing
that reacts to it.

\paragraph{Swarm (fan-out/fan-in).} One coordinator posts $n$ tasks under one
topic, each carrying a shard of the work, assigns them to a kind, and watches
the \emph{topic}. The Board Keeper allocates $n$ workers. Each worker posts its
shard \emph{done} with its output location. The coordinator wakes once per
completion, and on each wake it reads the board and asks whether the
termination condition is met---all shards done, or enough of them, or a
deadline passed. Two details make this work where a naive fan-out does not.
The coordinator's state lives on the board rather than in its head, so it can
count completions by reading rather than remembering; and partial failure is
ordinary, because a shard that fails is a task whose content says so and whose
re-posting is one more post.

\paragraph{Escalation to a human.} A task assigned to no agent, marked
\emph{waiting}, with content written for a person: what was decided, what is
blocked, and what answer would unblock it. The agent watches it and ends its
turn. A person moves it or writes into it from the board's kanban view, and
that post wakes the agent. The human is a participant, not a mechanism. The
reason this is a kernel rather than a feature is that the only thing that makes
it work is a content contract---write for someone who has not read your
conversation---and a content contract is precisely what a kernel is.

\begin{table}[t]
  \centering
  \small
  \setlength{\tabcolsep}{4pt}
  \begin{tabular}{@{}lll@{}}
    \toprule
    Kernel & Roles & Waits on \\
    \midrule
    Pipeline wait & agent & the Board Keeper \\
    Delegation    & asker, worker & one task \\
    Swarm         & coordinator, workers & a topic \\
    Escalation    & agent, person & one task \\
    Kanban        & planner, worker, reviewer & a topic \\
    \bottomrule
  \end{tabular}
  \caption{The kernel catalogue, smallest first. Every kernel is instructions
  in an agent's context; the board enforces none of them.}
  \label{kernels}
\end{table}

\section{What the Board Actually Guarantees} \label{properties}

A coordination substrate whose protocols are advisory invites an obvious
question: what is true no matter how badly the agents behave? The answer is a
short list, and we think stating it plainly is more useful than claiming more.

\paragraph{Guaranteed by the mechanism.}
\begin{compactitem}
  \item \textbf{G1, no self-wake.} An agent is never notified of its own post.
  No agent can spin by writing to the board.
  \item \textbf{G2, at most one notification per post.} An agent holding
  several watches that one post matches is woken once. Wake cost is bounded by
  posts, not by watches.
  \item \textbf{G3, no lost wakeup.} Setting a watch wakes the Board Keeper,
  which re-evaluates known events against the board. An event that completed
  before the watch was set is still delivered (\S\ref{keeper}).
  \item \textbf{G4, serialized keeper decisions.} All Board Keeper turns for a
  board run on one thread, so allocation and event-posting decisions are
  totally ordered.
  \item \textbf{G5, idempotent event delivery.} A task whose content already
  records an event is not posted to again for that event, so a waiter does not
  hear the same news twice.
  \item \textbf{G6, attribution and durability.} Every post appends its author
  to \emph{agents involved} and updates \emph{last updated}. The board is a
  durable, attributed record of who said what, readable by agents and people
  with the same tools.
\end{compactitem}

\paragraph{Not guaranteed; left to kernels.}
\begin{compactitem}
  \item \textbf{Mutual exclusion on assignment.} Nothing stops two agents
  claiming one task. The kanban kernel assigns that right to the planner alone;
  a kernel that forgets to is racy.
  \item \textbf{Status discipline.} Any agent may move any task to any status,
  including backwards. $\delta$ is advice.
  \item \textbf{Lost updates.} Replace-in-full means the last writer wins. Two
  agents posting to one task concurrently will lose one of the posts, and the
  board does not detect it. In practice tasks have one owner, which is a
  convention, not an invariant.
  \item \textbf{Termination.} Nothing guarantees an agent ever posts the task
  it was given. A worker that dies mid-task leaves it \emph{in progress}
  forever, and the asker waits forever.
  \item \textbf{Deadlock freedom.} Two agents can each wait on a task the other
  intends to post. Because wait conditions are prose, the board cannot build a
  wait-for graph and cannot detect the cycle structurally. This is the sharpest
  cost of the semi-semantic design, and we do not have a mechanism answer to
  it. What we have are mitigations: the board is visible, so a cycle is
  apparent to a person looking at the kanban; a \emph{done} task expires after a
  day, which bounds the lifetime of the record if not of the wait; and the
  Board Keeper, being an agent that reads the whole board, is in a position to
  notice and say so. Making it do that reliably is future work.
\end{compactitem}

\noindent
The honest summary is that the board guarantees the delivery properties and
almost none of the protocol properties. We think this is the right split for
the current state of the art, for the same reason that the board's wait
conditions are prose: the set of coordination patterns is not yet known, and a
mechanism that enforced today's patterns would be an obstacle to next
quarter's. We expect to migrate specific properties from the second list to the
first as particular kernels prove durable---assignment exclusivity is the
obvious first candidate, and it needs nothing more than a compare-and-set on
the assignee field.

\section{A Task from Request to Approval} \label{sequence}

\begin{figure*}[t]
\centering
\resizebox{\textwidth}{!}{%
\begin{tikzpicture}[x=1cm,y=1cm]
  \def\top{0}
  \def\bot{-14.3}
  \foreach \x/\n/\c in {0.9/{External\\agent}/cReq, 3.9/{Board}/black, 6.9/{Board\\Keeper}/cSys,
      9.9/{Worker}/cWrk, 12.9/{Pipeline}/cSys, 15.9/{Person}/cHum}{
    \node[box,draw=\c,fill=\c!12,minimum width=2.0cm] (h\x) at (\x,\top) {\n};
    \draw[life] (\x,\top-0.45) -- (\x,\bot);
  }
  \begin{scope}[on background layer]
    \fill[cReq!7]  (-0.6,-0.9) rectangle (17.5,-3.1);
    \fill[cInt!7]  (-0.6,-3.1) rectangle (17.5,-5.1);
    \fill[cWrk!7]  (-0.6,-5.1) rectangle (17.5,-9.2);
    \fill[cHum!7]  (-0.6,-9.2) rectangle (17.5,-11.3);
    \fill[cWrk!11] (-0.6,-11.3) rectangle (17.5,-13.7);
  \end{scope}
  \node[lbl,anchor=north west,font=\sffamily\tiny\bfseries] at (-0.55,-0.9) {1  REQUEST};
  \node[lbl,anchor=north west,font=\sffamily\tiny\bfseries] at (-0.55,-3.1) {2  ASSIGN};
  \node[lbl,anchor=north west,font=\sffamily\tiny\bfseries] at (-0.55,-5.1) {3  WORK AND WAIT};
  \node[lbl,anchor=north west,font=\sffamily\tiny\bfseries] at (-0.55,-9.2) {4  APPROVAL};
  \node[lbl,anchor=north west,font=\sffamily\tiny\bfseries] at (-0.55,-11.3) {5  REWORK AND CLOSE};
  \msg{-1.5}{0.9}{3.9}{post: task, brief, kind = worker}
  \node[box,draw=cReq,fill=white,minimum height=4mm] at (3.9,-1.95) {task \emph{planned}};
  \msg{-2.5}{0.9}{3.9}{watch task}
  \node[lbl,anchor=west] at (1.05,-2.9) {ends its turn; nothing of it is running};
  \msgd{-3.6}{3.9}{6.9}{watch on an unassigned kind}
  \node[box,draw=cSys,fill=white,minimum height=4mm] at (6.9,-4.15) {allocate worker};
  \msg{-4.7}{6.9}{9.9}{start on task}
  \msg{-4.95}{6.9}{3.9}{assign: kind $\to$ instance}
  \msg{-5.7}{9.9}{3.9}{claim: \emph{in progress}}
  \msg{-6.4}{9.9}{12.9}{start job}
  \msg{-6.65}{9.9}{3.9}{watch task, content says what to do on finish}
  \node[lbl,anchor=west] at (10.05,-7.05) {worker ends its turn};
  \msgd{-7.75}{12.9}{6.9}{job finished: pipeline wakes the Board Keeper}
  \msg{-8.3}{6.9}{3.9}{post event into the waiting task}
  \msgd{-8.8}{3.9}{9.9}{wake}
  \msg{-9.5}{9.9}{3.9}{write output; \emph{review}, assign to person}
  \msgd{-10.0}{3.9}{15.9}{wake: approval needed}
  \node[lbl,anchor=east,align=right] at (15.75,-10.55) {may be days later};
  \node[box,draw=cHum,fill=white,minimum height=4mm] at (15.9,-10.9) {reads the row};
  \msg{-11.7}{15.9}{3.9}{reject: write what is missing}
  \msgd{-12.15}{3.9}{9.9}{wake}
  \msg{-12.6}{9.9}{3.9}{revise; \emph{review}}
  \msg{-13.05}{15.9}{3.9}{approve: \emph{done}}
  \msgd{-13.5}{3.9}{0.9}{wake: result is ready}
\end{tikzpicture}}
\caption{One task from request to approval, composed from the kernels of
\S\ref{policy-kernels}. Solid arrows are posts to the board, dashed arrows are
wakeups. Dotted vertical lines are lifelines, and a participant is running only
where it sends an arrow or acts on a row; everywhere else it is blocked and
consuming nothing. The external agent is woken exactly once, when the task is
done. The person is woken only for the approval, and may answer hours or days
later.}
\label{sequence-fig}
\end{figure*}
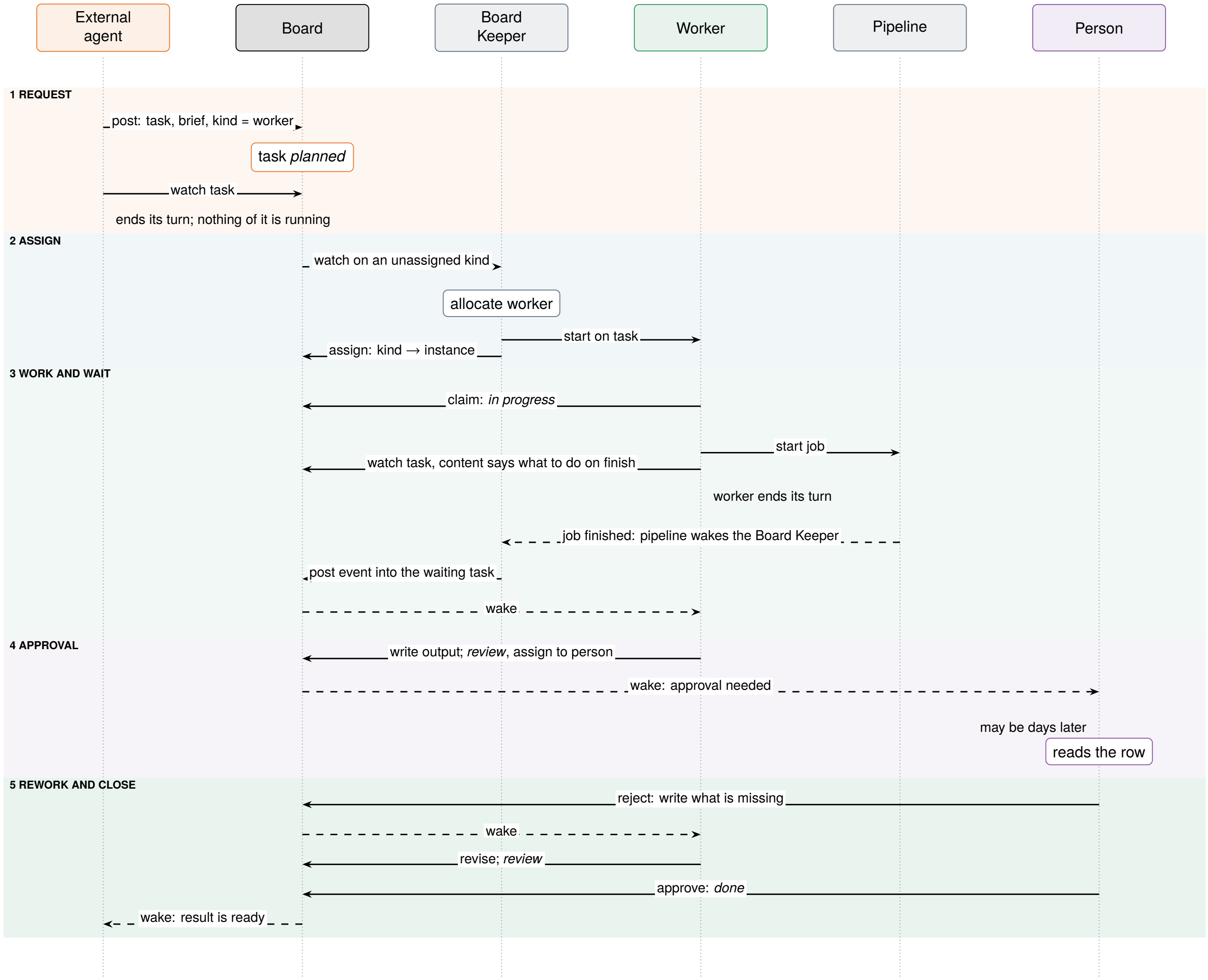

Figure~\ref{sequence-fig} follows one task end to end across every kind of
participant: a request from outside, assignment by kind, a worker that blocks
on a pipeline, an approval that goes to a person, and a rework round. It is a
composite of the kernels in \S\ref{policy-kernels}, not a trace; the next
section gives a real one.

The things to notice are all about who is awake. Between the first post and the
last, no participant runs for more than a few turns, and the longest stretches---the pipeline, the approval---have nobody resident
at all. The person
does not need to be at their desk when the approval post lands; the post is
durable, and their answer is the wakeup. The external agent never learns which
worker did the job, or that a pipeline was involved, or that the first draft
was sent back; it posted a row and was woken when the row said \emph{done}.

\section{A Trace from a Live Board} \label{trace}

Figure~\ref{trace-fig} is an abridged pair of rows from the board of a working
Corvic room, reproduced from the board table rather than reconstructed. Two
agents are involved and neither knew about the other when the first one
started.

The asking agent was a chat agent working on a document. It needed a LaTeX
document compiled, which needs a sandbox with a TeX distribution---a tool
catalog the chat agent does not have and cannot be given, because installing a
TeX distribution is not a thing a chat agent should be able to do to the room
it is talking in. It posted a task carrying the complete brief, assigned to the
\emph{kind} \textsf{computer use}, and watched it.

The Board Keeper allocated an agent of that kind---\textsf{sys/grackle/5426} in
the \emph{agents involved} list---started it on the task, and rewrote the
assignment from the kind to the instance, which is why \emph{assigned agents}
names \textsf{5426} in the final row. The worker ran for 5.6 minutes in
its own sandbox, making 43 model calls according to its trajectory. It fetched the
source files over the room's content API, installed a TeX tree, compiled, and
rendered 18 page images. It did not finish cleanly: its run ended in a
\texttt{RepeatedFormatError}, its reply file read ``The delegate stopped
without a reply'', and the build report and page notes the brief required were
never written. The board nevertheless shows the task as \emph{done}. The last
line of the task content is the Board Keeper's own writing: it names the
reply file and the feature view holding the worker's output directory, which is
the asking agent's handle on everything the worker produced, including the PDF
and log it left behind.

That post woke the asking agent. The worker had left no report, so the
asking agent evidently worked from the files it had left: it fixed two defects
in the source and posted a \emph{second} task---the second row in
Figure~\ref{trace-fig}---whose brief names the two regressions to check (a
multiply-defined citation and an empty bibliography entry) and, notably, passes
forward what the first worker had learned about which package archive actually
serves the document class. The second worker,
\textsf{sys/grackle/5427}, is a different agent with no memory of the first.
The knowledge travelled between them through the board.

Four things in this trace are worth drawing attention to. First, the
\emph{agents involved} column of both rows lists three parties in posting
order---the chat agent, the Board Keeper, and the worker---one of whom is the
system. The board records the collaboration, not just the result. Second, the
two workers ran for 5.6 and 5.4 minutes of model-timestamped activity, with
about 17 minutes between the end of the first and the start of the second;
during that gap the only participant doing anything was the asking agent. We did not
count the asking agent's turns. Third, the brief in the second task is visibly
better than the brief in the first, because it was written by an agent that had
seen what the first run produced. That is what iteration between agents looks
like when the medium is durable: not a conversation, but a sequence of
increasingly well-specified rows.

Fourth, and the reason we kept this trace and did not replace it with a
cleaner one, the first row is a failure that the board records as a success.
The pattern is not a one-off. Across all 16 delegations in this room, 13 ended
normally, two ended in a server error from the model provider, and one ended in
a format error. All three failures appear on the board as \emph{done}, and the
one that died on its first model call looks identical to one that completed 31.
The information existed---the exit status and the reply file said so---but it
reached the worker's output directory and not the task's status. We report the
count of 16 as a case study from one deployment run by the authors, mostly
LaTeX compile jobs, and not as a rate: it is the evidence for the termination
gap in \S\ref{properties}. The board does have a \emph{failed} status, and an
agent can write it. The defect is that a task a delegated worker reports into
is marked \emph{done} by the system when the worker stops, whatever the
worker's exit status was, so the status that exists is not the one the system
sets.

\begin{figure*}[t]
\centering
\footnotesize
\begin{minipage}{0.93\textwidth}
\hrule\vspace{3pt}
\begin{verbatim}
board_task_id    83570cb6e8fe5168b98ebfc4a5ff3a1d
topic            Compile SPUR paper v10
task_status      done
assigned_agents  [5426]
agents_involved  [chat-thread/4028/14540, board-keeper, grackle/5426]
last_updated     2026-10-01T01:00:20Z
task_content     Compile a LaTeX paper to PDF and report on the result.
                 THE SOURCE FILES  They live in a Corvic file table (feature
                 view ref 387016:artifact) ... 12 rows, each a file:
                 paper.tex plus eleven PNG figures ...  WHAT TO DO
                 paper.tex uses the acmart class. If acmart.cls is not in
                 your TeX tree, install it ...  Run pdflatex three times ...
                 [1400 words of brief omitted]
                 --- appended by the Board Keeper on completion: ---
                 Compile SPUR paper v10-668edcbe reported in
                 delegate-reply.md of feature view 387069:output, among the
                 files it left. Read it with view_files.

board_task_id    dbf7f65f8bdb50429c6dabc4a5c538a3
topic            Compile SPUR paper v10 fixed
task_status      done
assigned_agents  [5427]
agents_involved  [chat-thread/4028/14540, board-keeper, grackle/5427]
last_updated     2026-10-01T01:21:07Z
task_content     Compile a LaTeX paper to PDF ... This is a re-run of an
                 earlier successful build after two source corrections, so
                 the main job is confirming two specific defects are gone.
                 ... A previous agent compiled this successfully in a sandbox
                 like yours; the route that worked was installing TeX Live
                 packages into a local tree ... the tlnet archive serves the
                 packages (acmart, libertine, newtx, inconsolata, ncctools,
                 preprint, fontaxes, xstring, xkeyval and the kastrup binhex
                 dependency were all needed) ...
                 THE TWO REGRESSION CHECKS ...
                 [1500 words of brief omitted]
                 --- appended by the Board Keeper on completion: ---
                 Compile SPUR paper v10 fixed-d9dbdfde reported in
                 delegate-reply.md of feature view 387089:output, among the
                 files it left. Read it with view_files.
\end{verbatim}
\vspace{-6pt}\hrule
\end{minipage}
\caption{Two rows from a live Agent Board, abridged. One chat agent, two
sandboxed worker agents allocated by the Board Keeper, and the Board Keeper
itself. The second brief carries knowledge the first worker discovered; the
workers never met.}
\label{trace-fig}
\end{figure*}

\section{Agents Talking to Agents} \label{future}

We want to make a larger claim than ``this is a useful feature of our
harness'', and this section is where it belongs.

The current conversation about agent interoperability is overwhelmingly about
transport. MCP~\cite{mcp} standardizes how an agent reaches a tool;
A2A~\cite{a2a} standardizes how an agent reaches another agent. Both are
necessary and neither is sufficient, for the reason that dominates every system
in this paper: the hard part of multi-agent work is not the message, it is the
\emph{waiting}, and a protocol that defines an envelope has said nothing about
what happens during the four hours between the request and the answer.

A board answers that question in a way a protocol cannot, because a board is a
place rather than a path. Four properties follow from being a place, and we
think they are the ones that matter for agents talking to each other at scale.

The products of \S\ref{persistent-products} make this concrete and urgent. A
Muse agent lives on a Meta VM, a dot on an OpenAI cloud computer, a Grok Bot
on a SpaceXAI one. None can enter another's machine, and none should be able
to. When a user's dot needs something their company's internal agent
produces, there is no shared runtime to put that handoff in---and a
point-to-point message between them still leaves nobody holding a record of
what was owed. A board is the one artifact all three can touch without any of
them conceding anything about where they run.

\paragraph{The receiver need not exist.} A message must be addressed. A task
need only describe what kind of thing should do it. This is what lets work be
posted before anyone is available to do it, which is the normal case in any
system where capacity is elastic and the work is discovered rather than
planned. Contract Net~\cite{contractnet} understood this in 1980; the board's
assignment-by-kind is the same insight with the Board Keeper standing in for
the bidding round.

\paragraph{Non-agents are first-class.} A pipeline, a quota limit, a cron
schedule, and a person are all things an agent waits on, and none of them can
speak a protocol. On a board they do not have to: the Board Keeper writes on
their behalf, and the waiting agent cannot tell the difference. Every
coordination design that makes agents the only participants has to bolt the
rest of the world on afterwards.

\paragraph{The medium is the audit log.} Because every post is durable and
attributed, the record of what the agents agreed is the same object as the
mechanism by which they agreed it. There is no reconstruction step, no tracing
infrastructure to correlate, and no gap between what happened and what was
logged. For agents acting across organizational boundaries---where the question
``who authorized this and when'' is not optional---this is probably the
decisive property.

\paragraph{Prose survives version skew.} Two agents from different vendors,
with different tool catalogs and different models, cannot be expected to share
a schema for the work. They can be expected to read English. A semi-semantic
record degrades gracefully across that gap in a way a typed message does not:
the fields that must match exactly are few and dull, and everything
interesting is in a medium both parties already handle.

\noindent
Put together, these suggest the unit of agent interoperability should be a
shared, durable, prose-carrying board with a small exact routing core, and that
wire protocols should be how you reach the board rather than how you reach each
other. We are some distance from being able to claim that: the board described
here lives inside one room of one harness, trust is uniform within it, and
every participant is Corvic's. The obvious next questions---boards that span
trust boundaries, what a task looks like when the poster and the worker belong
to different organizations, how a board is partitioned without losing the
property that made it useful---are open, and we would rather name them than
pretend the design already answers them.

\section{Related Work}

\paragraph{Blackboard architectures.} Hearsay-II~\cite{hearsay2} and
BB1~\cite{hayesroth1985} established the shared-structure model: independent
knowledge sources reading and writing a common blackboard, with a control
component deciding who runs. The Agent Board is a blackboard with two
additions. The first is blocking: a knowledge source in Hearsay-II is invoked
by the scheduler, whereas an agent on the board registers a wait and is woken.
The second is that the control component is itself a language model, which is
what makes the wait conditions expressible in prose. Corkill's
retrospective~\cite{corkill1991} notes that blackboard control was the part
that never generalized; we would say the same and offer the Board Keeper as a
reason to revisit it.

\paragraph{Agent communication and task allocation.} Contract
Net~\cite{contractnet} is the ancestor of assignment-by-kind: announce a task,
let capable nodes bid, award it. KQML~\cite{kqml} and FIPA-ACL~\cite{fipa}
standardized speech-act message types between agents. These are protocols for
\emph{messages}; the board is a protocol for \emph{records}, and the difference
shows up precisely in the long-running case, where what you want persisted is
the obligation rather than the utterance.

\paragraph{Coordination languages.} Linda's tuple space~\cite{linda} is the
closest structural ancestor: a shared associative store where processes
coordinate by posting and taking tuples, decoupled in both space and time. The
board is a tuple space whose matching is semantic rather than structural, whose
tuples are durable and attributed rather than consumed, and whose blocking
read is a watch. The actor model~\cite{actors} and
Erlang/OTP~\cite{armstrong2003} take the opposite route---mailboxes and
supervision rather than shared structure---and OTP's supervision trees are a
better-engineered answer to the termination problem we admit to in
\S\ref{properties}.

\paragraph{Operating systems.} The wait-queue analogy runs through this paper.
Monitors and condition variables~\cite{hoare1974} are the direct antecedent of
watch-and-block, including the sleep--wakeup race our wake-on-watch rule
addresses; Dijkstra's semaphores~\cite{dijkstra1968} are the origin of the
problem. Futures~\cite{futures} are what a delegation task is when the wait is
synchronous. The level-triggered reconciliation we describe in \S\ref{keeper}
is the control-loop discipline of Borg and Kubernetes~\cite{borg,burns2016}.

\paragraph{Workflow and durable execution.} Airflow~\cite{airflow} and
Temporal~\cite{temporal} solve long-running coordination for programs, and
Temporal in particular solves the free-waiting problem exactly. The difference
is who writes the graph: these systems require it before the run, and the cases
that motivated the board are the ones where the agents write it during.
MapReduce~\cite{mapreduce} is the ancestor of the swarm kernel.

\paragraph{Persistent agent products.} Grok Bot~\cite{grokbot,grokbot-design},
Muse~\cite{muse,muse-security} and dots~\cite{dots,dots-vb} are the
industrial statement of the problem this paper addresses, and
\S\ref{persistent-products} is our full account of them. Each gives an agent
residency---a machine, a browser, and a life beyond one conversation---and
each draws its control boundary at that machine. The board is the layer above:
what these systems need once there is more than one agent, once the agent
spends most of its life waiting, and once the other agent belongs to someone
else.

\paragraph{LLM multi-agent systems.} AutoGen~\cite{autogen} organizes agents as
a group conversation; MetaGPT~\cite{metagpt} assigns them software-engineering
roles with standardized documents between them; Generative
Agents~\cite{generativeagents} gives each agent a memory stream and lets them
interact in a world. These are closest in spirit, and the gap is uniform: the
coordination substrate is the conversation, so an agent that must wait either
holds the floor or drops out. LangGraph~\cite{langgraph} and
CrewAI~\cite{crewai} are the engineering counterparts, and are orchestrators in
the sense of \S\ref{motivation}. Hermes Agent~\cite{hermes-kanban} is the
closest existing design to the kanban kernel, and the comparison in
\S\ref{policy-kernels} is the sharpest statement of the difference: a timed
dispatcher with explicit parent--child links, versus an event-woken planner
reading dependencies out of prose.

\section{WaitBench: A Benchmark for Long-Running Coordination}
\label{benchmark}

Section~\ref{properties} states what the board guarantees, and the
limitations say we have not measured it. Measuring only our own system would
still leave the field without a way to compare harnesses on the thing this
paper is about. We therefore propose a benchmark that any harness can run. The
working name is \emph{WaitBench}. This section is a design. We ran pilots of
three of its scenario families, reported in \S\ref{pilot-e1},
\S\ref{pilot-e3} and \S\ref{pilot-e2}, and \S\ref{hypotheses} states what we
expected as predictions the benchmark could falsify, with the pilot outcome
beside each.

\subsection{What existing benchmarks measure}

Most long-horizon benchmarks ask how long a single agent stays competent.
Table~\ref{bench-landscape} lists the closest ones we found. The search was
limited, two rounds of queries, and is not a survey.

\begin{table}[t]
  \centering
  \small
  \begin{tabular}{@{}p{0.27\columnwidth}p{0.62\columnwidth}@{}}
    \toprule
    Benchmark & What it measures \\
    \midrule
    METR time horizon~\cite{mettimehorizon} & Task length, as human-expert time, at which a model succeeds with a given reliability. \\
    Long-Horizon-Terminal-Bench~\cite{lhtb} & 46 terminal tasks with partial-credit grading; reported runs average 239 episodes, 9.8M tokens and 88.9 minutes. \\
    Vending-Bench~\cite{vendingbench} & Long-term coherence of one agent running a simulated business over runs above 20M tokens. \\
    ParaCook~\cite{paracook} & Time efficiency of scheduling and coordinating agents in a parallel cooking simulation. \\
    Gaia2~\cite{gaia2} & Agents in asynchronous environments with an event clock, temporal reasoning and multi-agent collaboration; a wait tool lets hours of simulated time run in minutes. \\
    Handoff Debt~\cite{handoffdebt} & Rediscovery cost when a coding agent takes over an interrupted task from a predecessor. \\
    \bottomrule
  \end{tabular}
  \caption{Neighbouring benchmarks and the object each one measures.}
  \label{bench-landscape}
\end{table}

Gaia2 is the nearest neighbour, and WaitBench borrows its central idea: an
event-driven simulated clock, so that a day of waiting costs minutes of
wall-clock time. The difference is the object under test. In each of these
benchmarks the agent or model is what varies and the harness around it is held
fixed. None of them asks what the coordination substrate costs while agents
wait, whether a wakeup is delivered once and only once, or what becomes of an
obligation when its worker dies. Handoff Debt is the exception in spirit, since
it measures a handoff, but it varies what the predecessor leaves behind rather
than the substrate that carries it.

\subsection{Design}

The principle is to \emph{hold the model fixed and vary the harness}. Every
run uses the same model, the same tasks and the same seed, and differs only in
how the harness lets agents wait, wake, hand off and fail.

A harness under test needs to expose one small contract: an agent can end its
turn with a registered wait, and the harness resumes it when the event arrives.
A harness with no such primitive is still testable. It must hold a call open or
poll, and those are the two baselines the benchmark scores it against.

\begin{figure}[t]
\centering
\resizebox{\columnwidth}{!}{%
\begin{tikzpicture}[x=1cm,y=1cm]
  \node[box,draw=cInt,fill=cInt!12,minimum width=3.2cm,minimum height=3.6cm] (h) at (0,0) {};
  \node[font=\sffamily\scriptsize\bfseries] at (0,1.5) {Harness under test};
  \node[box,draw=cInt,fill=white,minimum width=2.6cm] at (0,0.6) {agents, same model};
  \node[box,draw=cInt,fill=white,minimum width=2.6cm] at (0,-0.3) {wait and wake};
  \node[box,draw=cInt,fill=white,minimum width=2.6cm] at (0,-1.2) {record of obligations};
  \node[box,draw=cSys,fill=cSys!12,minimum width=3.6cm,minimum height=3.6cm] (d) at (6.2,0) {};
  \node[font=\sffamily\scriptsize\bfseries] at (6.2,1.5) {Benchmark driver};
  \node[box,draw=cSys,fill=white,minimum width=3.0cm] at (6.2,0.75) {simulated clock};
  \node[box,draw=cSys,fill=white,minimum width=3.0cm] at (6.2,0.0) {event and fault injector};
  \node[box,draw=cHum,fill=white,minimum width=3.0cm] at (6.2,-0.75) {simulated person};
  \node[box,draw=cWrk,fill=white,minimum width=3.0cm] at (6.2,-1.5) {verifier and meter};
  \draw[msg] (1.75,0.5) -- node[lbl,above] {posts, waits} (4.35,0.5);
  \draw[msgd] (4.35,-0.5) -- node[lbl,below] {events, wakes} (1.75,-0.5);
\end{tikzpicture}}
\caption{WaitBench separates the harness under test from a driver that owns
time. The driver injects events, faults and approvals on a simulated clock,
checks end states, and meters model calls, tokens and resident compute.}
\label{bench-arch}
\end{figure}
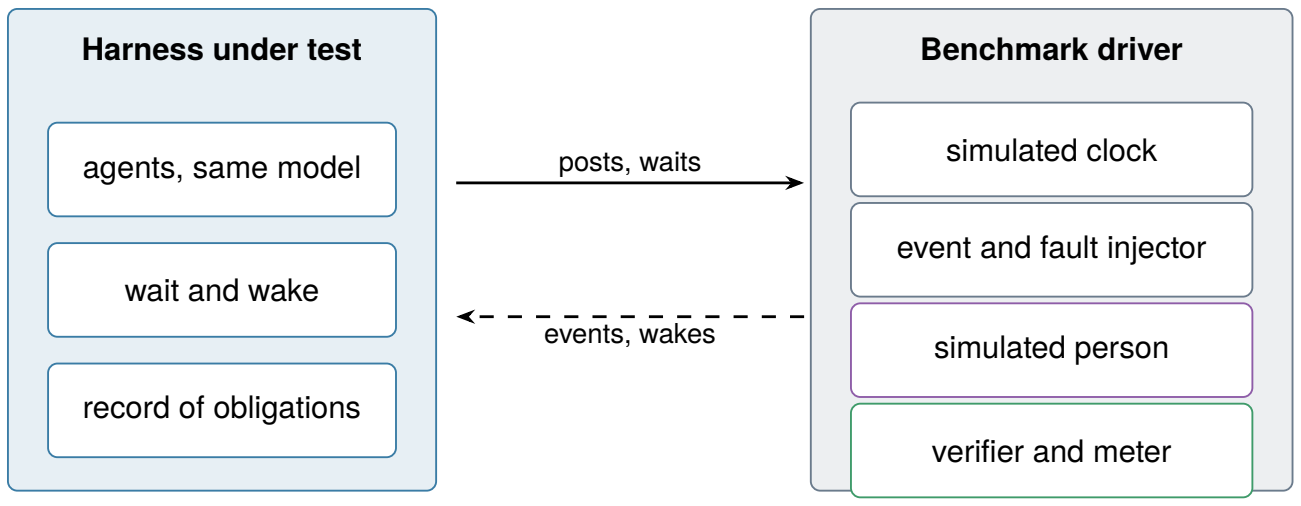

The driver supplies four things. A \emph{simulated clock} advances from event
to event, as in Gaia2. An \emph{injector} fires events at scheduled times,
including late, early and duplicate ones, and injects faults. A \emph{simulated
person} answers approval requests after a drawn latency and accepts or rejects
by a fixed policy. A \emph{verifier} checks the end state of each task against
a specification, so success never depends on a model grading itself. A meter
records model calls, tokens, and resident compute, meaning agent-seconds during
which a process or machine is held.

\subsection{Scenario families}

Table~\ref{bench-scenarios} lists nine scenario families. Each isolates one
property, so a harness's profile shows where it is strong rather than
collapsing to one score.

\begin{table*}[t]
  \centering
  \small
  \begin{tabular}{@{}l p{0.13\textwidth} p{0.40\textwidth} p{0.30\textwidth}@{}}
    \toprule
    ID & Family & What the driver injects & Primary measures \\
    \midrule
    S1 & Long wait & A job that finishes after $T \in \{10\,\mathrm{s}, 10\,\mathrm{min}, 1\,\mathrm{h}, 1\,\mathrm{day}\}$ of simulated time & Model calls, tokens and resident seconds as a function of $T$ \\
    S2 & Race & The event fires before, during, and after the watch is set & Missed-wake rate \\
    S3 & Crowd & $N$ concurrent waiters with prose conditions, $M$ events, and decoy events that answer no one & Wake precision and recall; duplicate wakes \\
    S4 & Handoff & A fresh agent resumes with the task record only, with full history, and with nothing & Success, turns and tokens to resume \\
    S5 & Approval & A person with a latency distribution who rejects the first draft with probability $p$ & Rework rounds; wakes per task; end-to-end time \\
    S6 & Failure & A worker dies, a pipeline fails, a model refuses, or an awaited event never arrives & Silent-stall rate; time to surface \\
    S7 & Swarm & $n$ shards of which $k$ fail & Completion detected; correct partial result \\
    S8 & Scale & From 1 to $10^4$ blocked agents & Cost per blocked agent-hour; wake latency \\
    S9 & Foreign poster & A requester that shares no runtime, file system or account with the worker & Task success with no shared state \\
    \bottomrule
  \end{tabular}
  \caption{WaitBench scenario families. S1--S3 test the wait primitive, S4--S5
  the continuation and the human, S6--S7 failure, and S8--S9 scale and
  boundaries.}
  \label{bench-scenarios}
\end{table*}

\subsection{Metrics}

\begin{compactitem}
  \item \textbf{Wait cost.} Model calls, tokens and dollars per wait, reported
  against $T$. This is the curve of \S\ref{blocking}, measured.
  \item \textbf{Resident cost.} Agent-seconds of held compute, reported
  separately from model cost. It separates residency from blocking, which
  \S\ref{persistent-products} argues are different things.
  \item \textbf{Wake quality.} Missed wakes, false wakes, duplicate wakes, and
  event-to-resumption latency on the simulated clock.
  \item \textbf{Handoff tax.} Extra turns and tokens a fresh agent needs to
  resume, relative to an agent that holds the full history.
  \item \textbf{Silent-stall rate.} The fraction of injected failures that are
  not surfaced anywhere a person or agent would see within a stated deadline.
  \item \textbf{Correctness.} Verifier pass rate on the end state.
\end{compactitem}

\subsection{Protocol and reporting}

Every scenario runs under a fixed seed set, with repeated runs per seed. A
report states the model, the number of runs, and an interval for each rate, and
it gives the denominator behind every percentage. Rates from fewer than about
100 runs, and means from fewer than about 30, are flagged as small samples.
Each submission carries a \emph{harness card}: what stays resident while an
agent waits, what is durable across a restart, and how a wakeup is delivered.
The card exists so that a reader can tell a harness that blocks from one that
merely polls cheaply.

We recommend that results for a given harness be produced by someone other
than its authors. The reference implementations of the poll and hold baselines
are the ones a reviewer would otherwise have to trust us on.

\subsection{What we expect}
\label{hypotheses}

These are predictions, not findings.

\begin{compactitem}
  \item \textbf{H1.} In S1, blocking cost is flat in $T$, polling cost grows
  with $T$, and holding the call fails beyond a harness-specific bound, as
  \S\ref{blocking} derives. \emph{Pilot result (\S\ref{pilot-e1}):} supported
  for flatness and for polling growth. It also showed that an external watcher
  is cheaper than the board, a prediction we had not made.
  \item \textbf{H2.} In S2, a harness that re-evaluates known events when a
  watch is set misses no wakeups because of timing. Any misses it has come from
  matching error, which S3 measures separately.
  \item \textbf{H3.} In S4, resuming from a task record written for that
  purpose costs fewer turns and tokens than rereading history. Handoff
  Debt~\cite{handoffdebt} reports a consistent efficiency gain of this kind for
  coding agents; S4 tests whether it holds for coordination records.
  \emph{Pilot result (\S\ref{pilot-e2}):} supported only for a note that lists
  the needed facts. A note the agent wrote on its own failed at 19 of 20 runs
  at $H{=}200$.
  \item \textbf{H4.} In S6, the board alone has a non-zero silent-stall rate,
  because \S\ref{properties} says it cannot detect a dead worker. A deadline
  kernel should lower it. We list this as a prediction against our own system
  deliberately.
\end{compactitem}

\subsection{Threats to validity}

A simulated clock and simulated person are cheaper and more repeatable than
real ones, and they may hide behaviour that only appears with real latency or
real people. Interpretive matching makes S3 depend on the model that does the
matching, so S3 results describe a harness--model pair. Any published scenario
set can be overfit. We would hold out a private split. Finally, the authors of
this paper built the reference harness, which is a reason to treat their runs
as a baseline and not as a verdict.

\paragraph{Status.} WaitBench is a proposal, and this paper does not promise a
release of its driver, scenarios or baselines. The three experiments of
\S\ref{pilot-e1}, \S\ref{pilot-e3} and \S\ref{pilot-e2} are pilots written
against the design in this section: they correspond to scenario families S1, S3
and S4, they use a single model, and none of them was run on Corvic's
production board. S2 and S5--S9 have not been run.

\section{Limitations}

We have described a mechanism and argued for it, and we ran three pilots
(\S\ref{pilot-e1}, \S\ref{pilot-e3}, \S\ref{pilot-e2}). They are small
simulations with one model and synthetic tasks, none of them ran on Corvic's
production board, and the paper should be read accordingly. \S\ref{benchmark}
proposes how a full evaluation should go, and the predictions there are the
ones to hold us to. Three limitations are structural rather than incidental.

\emph{Interpretation is not matching.} The Board Keeper decides by reading, so
it can be wrong: it can post an event into a task that was not waiting for it,
costing a turn, or fail to post one into a task that was, costing a hang. The
simulation of \S\ref{pilot-e3} measured both on easy boards: no wrong wakes in
250 positive and 50 negative trials, and 9 missed wakes in 250 (3.6\%). We have
no measurement for hard boards or for the production Keeper. A wrong wake is
cheap and a missed wake is not, which argues for tuning the Board Keeper toward
over-delivery, and G5 (idempotent delivery) is what makes over-delivery
survivable.

\emph{Advisory protocols are advisory.} Everything in \S\ref{policy-kernels} is
instructions, and an agent that ignores them corrupts the board for everyone
sharing it. The blast radius of a badly-behaved agent is the whole board.

\emph{Prose is unbounded.} Nothing limits how much an agent writes into a task,
and a task read by the Board Keeper on every event is a cost that grows with
what everyone has written into it. We expect this to be the first scaling wall
and we have not hit it yet.

\emph{What is not yet measured.} Four things a systems venue would ask for
are absent. We have not run any experiment on Corvic's production board, so
the cost of its Board Keeper, its queueing under load, and the resident
compute of its workers are unmeasured. We have not measured wake amplification
on a busy board, or how often agents violate their kernel's transition
relation. We have not tested boards of hundreds of waiters, or ambiguous
wait conditions. And finished tasks leave the board after a day, so measuring
these on a live deployment needs run traces and agent histories rather than the
board table; \S\ref{benchmark} proposes a harness for collecting them.

\section{Conclusion}

The Agent Board is a shared table of semi-semantic tasks, a blocking primitive
whose cost does not grow while it waits, and a system agent that carries the rest of
the harness onto the board and allocates agents for work that names no one.
Those three things together let an agent start work that outlives it, hand work
to an agent that does not yet exist, wait for a pipeline or a person for an
arbitrary length of time, and resume from a note it wrote itself rather than
from a history it must re-read.

We have argued that this is best understood as operating-system scheduling with
one substitution---a language model where the exact matcher used to be---and
that the substitution is what buys the expressiveness and what costs the
guarantees. The resulting system enforces delivery and leaves protocol to
convention, which we believe is the correct trade while the space of
coordination patterns is still being discovered.

We have also proposed a benchmark, WaitBench (\S\ref{benchmark}), because the
questions here are about the harness, and the existing long-horizon benchmarks
vary the model while holding the harness fixed. Our pilots give a first,
honest answer and a qualified one. The board's wait cost is flat and naive
polling's is not, but a plain external watcher is cheaper still, so the board's
value is in matching prose conditions, in the shared record, and in allocating
workers, not in cheaper waiting. The Keeper matched prose conditions well on
easy boards but missed 3.6\% of wakes. And a note replaces history only when it
says what the resumer needs. Whether wakeups are delivered once at scale, and
what happens when a worker dies, remain to be measured, and nobody should have
to take our word for them.

The larger claim is that this shape, and not a wire protocol, is what agents
talking to each other will need: a durable place where work is described, owed,
and recorded, which non-agents can be represented on, which humans can read,
and whose medium is the one thing every agent already understands.

2026 was the year the industry gave agents somewhere to live. Muse, dots and
Grok Bot each decided, separately, that an agent worth the name has to
outlast the conversation, and each bought that with a machine. We think the
next thing they discover is that a machine is a room with one chair in it. The
agents are about to need somewhere to meet: durable, legible, open to whoever
has work to hand over, and indifferent to whose cloud anybody is running in.
Teams worked this out some time ago and put a board on the wall. We are
arguing for the same thing, with the standup automated and the waiting made
cheap.

\bibliographystyle{plainnat}
\bibliography{agent_board}

\end{document}